\documentclass[twoside]{article}

\usepackage{microtype}
\usepackage{subcaption}
\usepackage{graphicx}
\usepackage{booktabs} 

\usepackage{pgfplots}
\pgfplotsset{compat=1.15}
\usepackage{mathrsfs}
\usetikzlibrary{arrows}

\usepackage{amsmath}
\usepackage{amssymb}
\usepackage{mathtools}
\usepackage{amsthm}
\usepackage{tikz}

\theoremstyle{plain}

\theoremstyle{definition}

\theoremstyle{remark}

\usepackage[accepted]{aistats2024}
\usepackage[round]{natbib}

\usepackage{hyperref}

\begin{document}

%
\runningtitle{Fixed-kinetic Neural Hamiltonian Flows}

%

\twocolumn[

\aistatstitle{Fixed-kinetic Neural Hamiltonian Flows for \\
enhanced interpretability and reduced complexity}

\runningauthor{Vincent Souveton, Arnaud Guillin, Jens Jasche, Guilhem Lavaux, Manon Michel}






\aistatsauthor{Vincent Souveton \\ LMBP, CNRS,\\ Université Clermont-Auvergne \And Arnaud Guillin \\ LMBP, IUF,\\ Université Clermont-Auvergne\\ \And Jens Jasche \\ Stockholm University \AND Guilhem Lavaux \\ IAP, CNRS, Sorbonne Université \And Manon Michel \\LMBP, CNRS, Université Clermont-Auvergne}
\aistatsaddress{}]

\begin{abstract}
  Normalizing Flows (NF) are Generative models which transform a simple prior distribution into the desired target. They however require the design of an invertible mapping whose Jacobian determinant has to be computable. Recently introduced, Neural Hamiltonian Flows (NHF) are Hamiltonian dynamics-based flows, which are continuous, volume-preserving and invertible and thus make for natural candidates for robust NF architectures. In particular, their similarity to classical Mechanics could lead to easier interpretability of the learned mapping. In this paper, we show that the current NHF architecture may still pose a challenge to interpretability. Inspired by Physics, we introduce a fixed-kinetic energy version of the model. This approach improves interpretability and robustness while requiring fewer parameters than the original model. We illustrate that on a 2D Gaussian mixture and on the MNIST and Fashion-MNIST datasets. Finally, we show how to adapt NHF to the context of Bayesian inference and illustrate the method on an example from cosmology.
\end{abstract}

\section{INTRODUCTION}

Generative models are now under growing interest regarding sampling high-dimensional probability distributions with applications from molecular biology \citep{lopez2020enhancing} to cosmology \citep{rodriguez2018fast} or medical science \citep{frazer2021disease}. Traditional architectures like Generative Adversarial Networks \citep{GANs} have shown impressive results in image generation but since their adversarial loss seeks a saddle point rather than a local minimum, GANs are notoriously hard to train and may suffer from mode-collapse \citep{2017arXiv171204086L,2017arXiv170104862A,2019arXiv190604848B}. More robust techniques like Normalizing Flows (NF) were thus developed \citep{originalNF, NFDensityModeling, NF_for_variational_inference}. NF consist in training a neural network to map a simple prior distribution onto the desired target through a chain of invertible transformations. They come with interesting characteristics, such as stability and correctness, see for example \cite{reviewNF}. The main limitation comes from the design of an invertible function for the mapping. In particular, computing the Jacobian determinant in the change of variable formula may be costly. Also, explainability is now a growing concern within the community \citep{interpretability}, in particular regarding applications in natural science, as the transformation learned by NF models is commonly hard to interpret.

First motivated by the mitigation of the Jacobian computation limitation, Neural Hamiltonian Flows (NHF) \citep{Toth2020Hamiltonian} are NF models that use Hamiltonian transformations. Indeed, in classical Newtonian mechanics, the Hamiltonian of a system, composed of a kinetic and a potential energy terms, sets its dynamical evolution, which is reversible and has a Jacobian determinant equal to one. They have exhibited performance similar to RealNVPs in sampling some 2D distributions \citep{Toth2020Hamiltonian}. Furthermore, being Physics-driven models, they are expected to enhance interpretability and it is furthermore straightforward to exploit the Hamiltonian properties to include some invariance under symmetrical transformations \citep{EquivariantHF}. However, the NHF architecture is made of four neural networks black-boxes that may render difficult the interpretation of the learned dynamics and energies. In particular, the learnt potential energy is not guaranteed to correspond to the corresponding physical potential energy of the data, when writing their probability distribution as a Boltzmann one. Even if it was numerically shown to transfer multimodality from the target distribution to the potential energy in some 2D cases \citep{Toth2020Hamiltonian}, this property is not ensured. 

We focus here precisely on the transfer of the negative logarithm of the target distribution into the learned potential, which should be the case in any physical system. This leads to propose a fixed-kinetic version of NHF (FK-NHF), where momenta follow a Gaussian distribution, i.e. its classical mechanics formulation. We discuss the impact of the hyperparameters on such transfer and study its overall robustness towards the choice of the numerical integration scheme and of the prior distributions. This work provides the following five main contributions: 
    \begin{itemize}
        \item We introduce a FK-NHF which, thanks to the Hamiltonian evolution, enhances the interpretability of the model at a cheaper computational price, compared to the standard multilayer perceptron version (MLPK-NHF). 
        \item We analyze the effect of multiple parameters of the architecture on sampling a 2D multimodal distribution and show that NHF is robust to the hyper-parameters choice, especially FK-NHF.
        \item We show that the choice of prior has an influence on the learned dynamics. FK-NHF allows better robustness to the choice of prior distribution. 
        \item We evaluate the sampling performance of NHF models on high-dimensional image generation problems and compare them to a classical RealNVP.
        \item Finally, aside from Generative modeling, flow-based models are relevant for inference \citep{NF_for_variational_inference, ConditionalNF}. We test a framework for Bayesian inference using NHF and present numerical experiments for inferring cosmological parameters from astronomical observations. The methodology we propose is inspired by Boltzmann generators \citep{boltzmannGenerators}.
    \end{itemize}

The manuscript is organized as follows: Section \ref{sec:RelatedWorks}  presents and reviews related works. In Section \ref{sec:NHF}, we describe the theoretical framework and the practical implementation of NHF. In Section \ref{sec:fixedKineticNHF}, we discuss the choice of models for the kinetic energy and introduce FK-NHF for enhanced interpretability and reduced complexity. Section \ref{sec:RobustnessOptimisation} discusses the maximization of expressivity given a fixed computational budget, with tests on a 2D Gaussian mixture and analysis of the impact of the Leapfrog-hyperparameters and model complexity. We also show how the choice of base distribution affects the learned energies and thus the interpretability of the model. In Section~\ref{sec:HDresults}, we discuss the generative performance of NHF models in high-dimensional problems. Finally, in Section \ref{sec:HNFforBayes}, we adapt NHF for Bayesian inference and illustrate our method on a standard model from cosmology.

\section{RELATED WORKS}
\label{sec:RelatedWorks}

\textbf{Generative models.} Various architectures have been presented such as Generative Adversarial Networks \citep{GANs} or diffusion networks \citep{diffusionNetworks}. Here, we will focus on Normalizing Flows techniques \citep{originalNF, NFDensityModeling, NF_for_variational_inference} as a way of smoothly transforming a simple prior distribution into the target posterior. The FK-NHF we present can however be understood as some ODE counterpart of diffusion models where the transformation is governed by a Langevin SDE. Both transformations agree for one discrete time step and this situation is reminiscent of the one in sampling with HMC \citep{MassMatrixHMC} and Metropolis-adjusted Langevin algorithm \citep{mala}. The main limitation of diffusion models is that they usually require many iterations ($\sim 1000$ \citep{ho}) to produce good samples, while we show that FK-NHF only needs around $10$ leapfrog steps.

\textbf{Learning Hamiltonians.} Learning Hamiltonians, i.e. physical conserved quantities, is a first step towards a better understanding of the physical processes that govern the data generation. Multiple architectures are proposed, such as Hamiltonian Neural Networks  \citep{HNN} or Hamiltonian Generative Networks \citep{Toth2020Hamiltonian}. These methods parameterize the Hamiltonian with neural networks and come with useful properties such as exact reversibility and smoothness. They have inspired applications from domain translation \citep{hamiltonian_domain_translation} to fault-detection in industry \citep{hnn_faults_detection}. Notably, they can be combined with Markov-Chain Monte Carlo (MCMC) methods, for instance as proposals in the Hamiltonian Monte Carlo (HMC) algorithm \citep{DUANE1987216, bayesian_inference_lhnn}. We learn here artificial Hamiltonians for sampling and our goal is to extract the negative logarithm of the target distribution into the potential. 

\textbf{Neural ODE Flows.} Compared to the transformation in Neural ODE flow \citep{ODE}, the Hamiltonian ODE in NHF are volume-preserving, making for a cheaper log-likelihood computation, and can be integrated via symplectic integrators. We discuss here its robustness with respect to hyperparameters and choice of prior distribution. Also, we propose the alternative FK version of NHF to enhance interpretability while reducing the complexity of the model.

\textbf{Inference with NHF.} Traditional MCMC methods \citep{MCMCmethods} are very popular because they come with guarantees in terms of convergence and much progress has been made regarding their tuning \citep{NUTS, Stan}. NF architectures have also been proposed in this framework \citep{NF_for_variational_inference, ConditionalNF}. Here, we adapt NHF to sampling Bayesian posterior distributions by transforming the prior distribution into the posterior with no access to samples from the target. 

\textbf{Explainable AI.} XAI deals with the problem of understanding the decisions made by an Artificial Intelligence \citep{XAI}. Indeed, complex architectures made of multiple (deep) neural networks are often easier to train than to understand. Some solutions involve surrogate techniques \citep{surrogates}, local perturbations \citep{localPerturbations} or meta-explanations \citep{metaExplanations}. Including physical prior knowledge into neural networks may be another solution to understand the model \citep{physicsInformedNN, Toth2020Hamiltonian}. In this work, we build on that idea and try to make the model as explainable as possible by fixing its kinetic energy and thus enforcing classical Mechanics knowledge into the architecture.

\section{NORMALIZING FLOWS WITH HAMILTONIAN TRANSFORMATIONS}
\label{sec:NHF}

\subsection{Normalizing Flows}

Normalizing flows are generative models mapping a complex target distribution $\pi$ onto a known prior distribution $\pi_0$ which is easy to sample \citep{reviewNF}. This mapping is a series of smooth invertible transformations $\mathcal{T}_1,...,\mathcal{T}_L$. Once the model is trained, one can reverse the learned dynamics to generate samples from the target distribution starting from the prior. If $X = \mathcal{T}_L \circ ... \circ \mathcal{T}_1(Z)$, where $Z \sim \pi_0$, the density followed by $X$ reads $m(x) = \pi_0\left( \mathcal{T}^{-1}_1 \circ ... \circ \mathcal{T}^{-1}_L(x) \right) \times \prod_{k=1}^L \left| \det J_{\mathcal{T}_k^{-1}}(x) \right|.
$ The model parameters to optimize are denoted $\Theta$. The goal is to minimize the Kullback-Leibler divergence between the target distribution $\pi$ and the model distribution $m$ with respect to $\Theta$, i.e. minimizing:
{\footnotesize\begin{align*}
	&\mathcal{L}(\Theta) = \mathbf{E}_{\pi} \left[ \log \pi(X) - \log m(X;\Theta) \right] \\
	&= -\mathbf{E}_{\pi} \Bigg[ \log \pi_0 \left( \mathcal{T}_1^{-1} \circ ... \circ \mathcal{T}_L^{-1}(X;\Theta) \right) \\
 &\qquad\left.+ \sum_{k=1}^L \left| \det J_{\mathcal{T}_k^{-1}}(X;\Theta) \right|  \right] + C.
\end{align*}}
One can use samples from the target distribution in order to get a Monte Carlo estimation of the above loss and minimize it with gradient descent.

At this point, transformations are to some extent arbitrary. Any smooth invertible transformation is suited but the main computational cost comes from the Jacobian determinants. The first goal then is to reduce this computational cost, and the second one to enhance interpretability, by a proper choice of the transformation, and of its induced inverse.

\subsection{Neural Hamiltonian Flows}

\begin{figure}[!h]
    \centering\includegraphics[width=1.0\hsize]{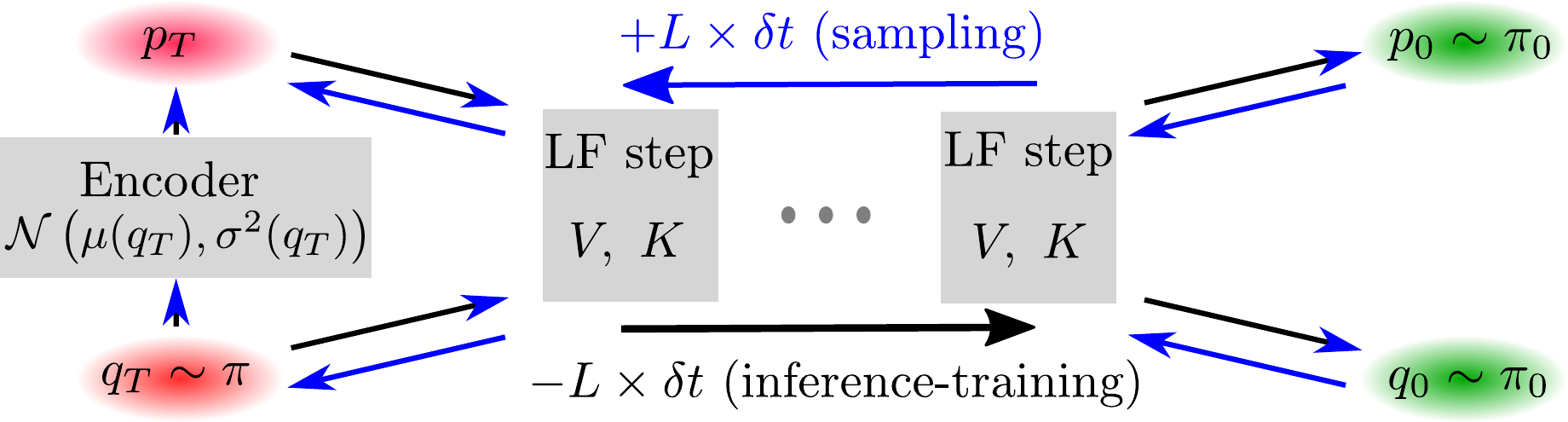}
    \caption{Schematic representation of the NHF architecture. In training, dataset samples are identified as generalized positions and the Encoder generates artificial generalized momenta. The system then evolves in the phase space following a discretized Hamiltonian flow. The resulting output must follow the prior distribution. The target data distribution can be sampled by inverting the learned dynamics.}
    \label{fig:architecture}
\end{figure}

To alleviate these issues, Neural Hamiltonian Flows \cite[NHF,][] {Toth2020Hamiltonian} is a NF technique that uses a series of Hamiltonian transformations as normalizing flows. In classical Mechanics, a system is fully described by its coordinates $(\mathbf{q},\mathbf{p})$ in phase-space. From that description, it is possible to define a scalar quantity called a Hamiltonian \citep{landau1982mechanics}. It can be seen as the total energy of the system and, in this paper, we make the assumption that it is written as the sum of a potential energy $V$, solely depending on the generalized positions $\mathbf{q}$, and a kinetic energy $K$, solely depending on the momenta $\mathbf{p}$. The system evolves in phase-space following Hamilton's equations that read:
\begin{equation}\label{Hamilton}
    \frac{d\mathbf{q}}{dt} = \frac{\partial H}{\partial \mathbf{p}}, \ \frac{d\mathbf{p}}{dt} = -\frac{\partial H}{\partial \mathbf{q}}.
\end{equation}
Hamiltonian transformations present at least two main advantages that make them suited for normalizing flows:
\begin{itemize}
    \item they are invertible by construction and inversion is easy by using a classical numerical integrator, i.e. just reversing the speed;
    \item their Jacobian determinant is equal to 1, removing the necessity to compute such determinant for each transformation. 
\end{itemize}

Numerically, the continuous solution can be approached by a symplectic, invertible and stable integrator as a Leapfrog:
\begin{equation}\label{leapfrogStep}
	\left\{	
	\begin{array}{rll}
	\mathbf{p}_{n+\frac{1}{2}} & = & \mathbf{p}_n - \nabla V (\mathbf{q}_n) \times \frac{\delta t}{2}, \\
	\mathbf{q}_{n+1} & = & \mathbf{q}_n + \nabla K(\mathbf{p}_{n+\frac{1}{2}}) \times \delta t, \\
	\mathbf{p}_{n+1} & = & \mathbf{p}_{n+\frac{1}{2}} - \nabla V (\mathbf{q}_{n+1}) \times \frac{\delta t}{2}.
	\end{array}
	\right.
\end{equation}

An illustration of a NHF model can be found in Figure~\ref{fig:architecture}. NHF is trained on a dataset consisting of realizations from the target distribution. To simulate Hamiltonian dynamics, one must extend the position space in which live the samples into the phase space, by adding artificial momenta: this is the role of the Encoder. The dynamics are integrated in phase-space with the Leapfrog integrator. More precisely, during training, NHF takes batches of $\mathbf{q}_T$ from the training dataset as inputs. For each $\mathbf{q}_T$, one $\mathbf{p}_T$ is drawn from a Gaussian distribution whose mean $\mu(\mathbf{q}_T)$ and deviation $\sigma(\mathbf{q}_T)$ depend on the $\mathbf{q}_T$. The resulting point in phase-space then evolves through a series of $L$ Leapfrog steps with integration timestep $-\delta t$. The outputs consist in the final position $\mathbf{q}_0$ and momenta $\mathbf{p}_0$, as well as the initial mean $\mu(\mathbf{q}_T)$, deviation $\sigma(\mathbf{q}_T)$ and $\mathbf{p}_T$ used in the loss computation. Once trained, one can easily define a sampling function that transforms $\mathbf{q}_0$, $\mathbf{p}_0$ into $\mathbf{q}_T$, by changing the sign of integration timestep and moving the system through the learned dynamics. An illustration of the architecture can be found in Figure~\ref{fig:architecture}.

Now regarding the training, following the previous notations, let  $f(.|\mathbf{q}_T)$ be the density of a normal distribution $\mathcal{N}(\mu(\mathbf{q}_T),\sigma(\mathbf{q}_T)^2)$, and $\mathcal{T}^{-1}$ the backward transformation of phase-space performed by NHF i.e. $\mathcal{T}^{-1}(\mathbf{q}_T,\mathbf{p}_T) = (\mathbf{q}_0,\mathbf{p}_0)$. Denote $\Pi_0$ the joint distribution of $\mathbf{q}_0$, $\mathbf{p}_0$. By adding artificial momenta $\mathbf{p}_T$ \citep{Toth2020Hamiltonian}, the distribution modeled by the NHF is $m(\mathbf{q}_T) = \int M(\mathbf{q}_T,\mathbf{p}_T) d\mathbf{p}_T = \int \Pi_0(\mathcal{T}^{-1}(\mathbf{q}_T,\mathbf{p}_T)) d\mathbf{p}_T$. This integral being intractable, one maximizes the following ELBO:
\begin{equation}\label{elbo}
    \mathcal{L}(\mathbf{q}_T) = \mathbf{E}_{f} \left[ \log \Pi_0(\mathcal{T}^{-1}(\mathbf{q}_T,\mathbf{p}_T)) - \log f(\mathbf{p}_T|\mathbf{q}_T) \right].
\end{equation}
    
This quantity is approximated via Monte Carlo integration.  Having learned the transformation, one can reverse the sign of timesteps and use the same potentials to transform the prior distribution into the target distribution. 

To summarize, the first part of the architecture consists of adding artificial momenta, as done by the Encoder, to simulate Hamiltonian dynamics. Here, $\mu$ and $\sigma$ are approximated by two neural networks. As for the Hamiltonian transformations, they are made by chaining Leapfrog steps. To do so, one must design the potential energy $V$ and the kinetic energy $K$ of the system. In \cite{Toth2020Hamiltonian}, each energy term is parameterized by a neural network. We will discuss this choice in the following section. For now, let us highlight that integrating Hamilton's equations with a symplectic numerical scheme provides flexibility. Most NF architectures rely on a careful architecture design rendering the computation of the Jacobian determinant easy \citep{RealNVP}. This is not the case with NHF since invertibility and volume-conservation are ensured by the use of a Leapfrog integrator and do not depend on the neural networks that are used to parameterize $\mu$, $\sigma$, $V$ and $K$.

\section{DESIGNING THE KINETIC ENERGY FOR NHF}
\label{sec:fixedKineticNHF}

 \textbf{MLPK-NHF.} If the kinetic energy is chosen to be a MLP \citep{Toth2020Hamiltonian}, then the model contains two black-boxes that are not easy to interpret \textit{a priori}, namely kinetic and potential energies $K$ and $V$. In particular, when sampling a multimodal distribution from an unimodal prior,  the learnt potential $V$ may not reflect the multimodal distribution.
 
 \textbf{FK-NHF.} By fixing the kinetic energy inside NHF, we gain interpretability on the learned flow by forcing the latter to obey some Physics principles. In this model, $K$ is no longer a MLP but a quadratic function $
K(\mathbf{p}) = \frac{1}{2}\mathbf{p}^T {\cal M}^{-1} \mathbf{p}$, with ${\cal M}$ a symmetric positive matrix. Starting from ($\mathbf{q}_0, \mathbf{p}_0)$ drawn from an unimodal prior distribution and imposing quadratic kinetic energy significantly reduces the possibilities for the potential energy to recover a multimodal $\mathbf{q}_T$. We indeed aim at enforcing these energies to be classical from a Physics perspective, i.e. making the learned kinetic energy to be of a quadratic form and the learned potential to be the negative logarithm of the target distribution $-\log \pi$ (or an approximation), as it is the case for diffusion models. It is noteworthy that, if the learnt potential is $-\log\pi$ and $K$ of a classical form, then any initial distribution of $\mathbf{q}_0$ can be mapped to the distribution $\pi$ of the $\mathbf{q}_T$ given the distribution of $\mathbf{p}_0$ is rich enough (e.g. a Normal law). Indeed, Hamiltonian dynamics with momenta refreshment yields an ergodic exploration of phase-space that leaves the canonical distribution ($\propto\exp(-\log\pi-K)$) invariant. Canonical distribution invariance uses volume-preservation in phase-space, i.e. Liouville's theorem, and ergodicity comes from momenta refreshment.

This is also at the basis of the HMC method \citep{DUANE1987216, MassMatrixHMC}. Therefore the FK variant can learn any distribution of $\mathbf{q}_T$. 
 
 Keeping track of the transformation dynamics, and the analogy with the familiar classical mechanics framework, makes possible the interpretability of the learned potential. Also, learning the energy landscape associated with the target distributions offers guarantees in terms of control of the discretization scheme, avoiding chaotic behaviour and making the model less sensitive to the choice of Leapfrog hyperparameters as we show in Section~\ref{sec:RobustnessOptimisation}. By doing so, both interpretability, sparsity and robustness are gained through FK-NHF. It is possible to create different versions of NHF by fixing its kinetic energy. One could, for instance, use a relativistic kinetic energy instead of a classical one. We now numerically show how the classical choice for kinetic energy yields an interpretable potential $V$ and such in a robust manner.

\section{INTERPRETABILITY AND ROBUSTNESS}
\label{sec:RobustnessOptimisation}

\begin{figure}[!h]
    \centering
    \includegraphics[width=1.0\hsize]{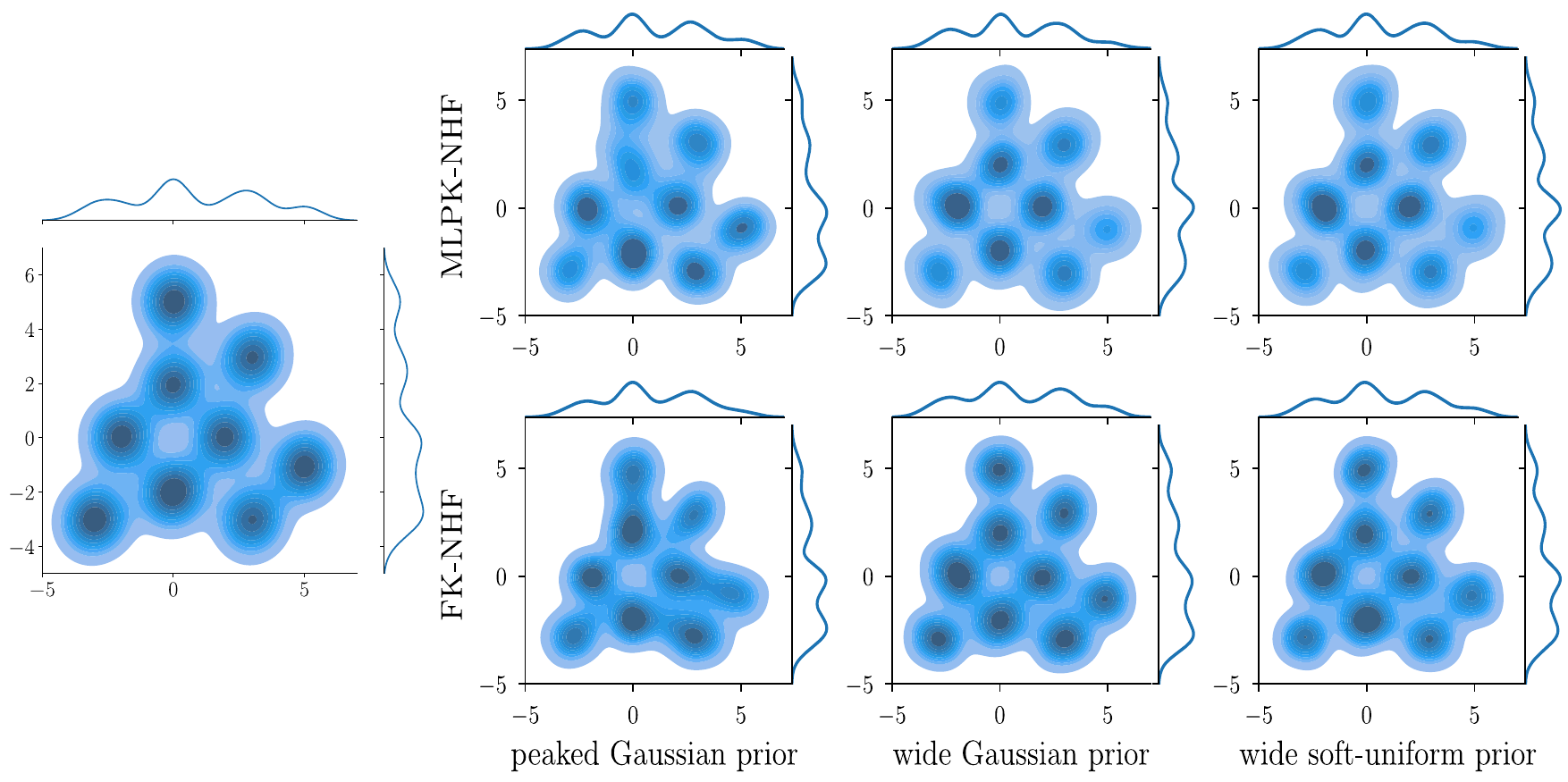}    
    \caption{Density estimation with its marginals of the target 2D Gaussian mixture (Left) and of the samples produced by MLPK-NHF (Right, Top) and
FK-NHF (Right, Bottom), with, from left to right, peaked Gaussian, wide Gaussian and wide soft-uniform prior. } 
    \label{fig:target}
\end{figure}

We present the results of numerical experiments for sampling a 2D Gaussian mixture (9 equally-weighted Gaussians with same covariance matrix $0.5^2 I_2$) (see Figure~\ref{fig:target}). Such an example, similarly studied in \cite{Toth2020Hamiltonian}, enables us to understand important aspects of NHF, like sensitivity to the choice of hyperparameters and, more importantly, interpretability. Also, traditional generative models like GANs may suffer from mode-collapse problems even in simple multimodal 2D settings \citep{GANmodeCollapse}, mode-collapses which were never observed with NHF experiments. Additional details about the models and hyperparameters choice can be found in Appendix~\ref{subsec:expDetails2D}.  A public version of the code is available on a Git repository \footnote{\url{https://plmlab.math.cnrs.fr/stoch-algo-phys/generative-models/fixed-kinetic-NHF/}}.

\begin{figure}[!h]
\centering
\includegraphics[width=1.0\hsize]{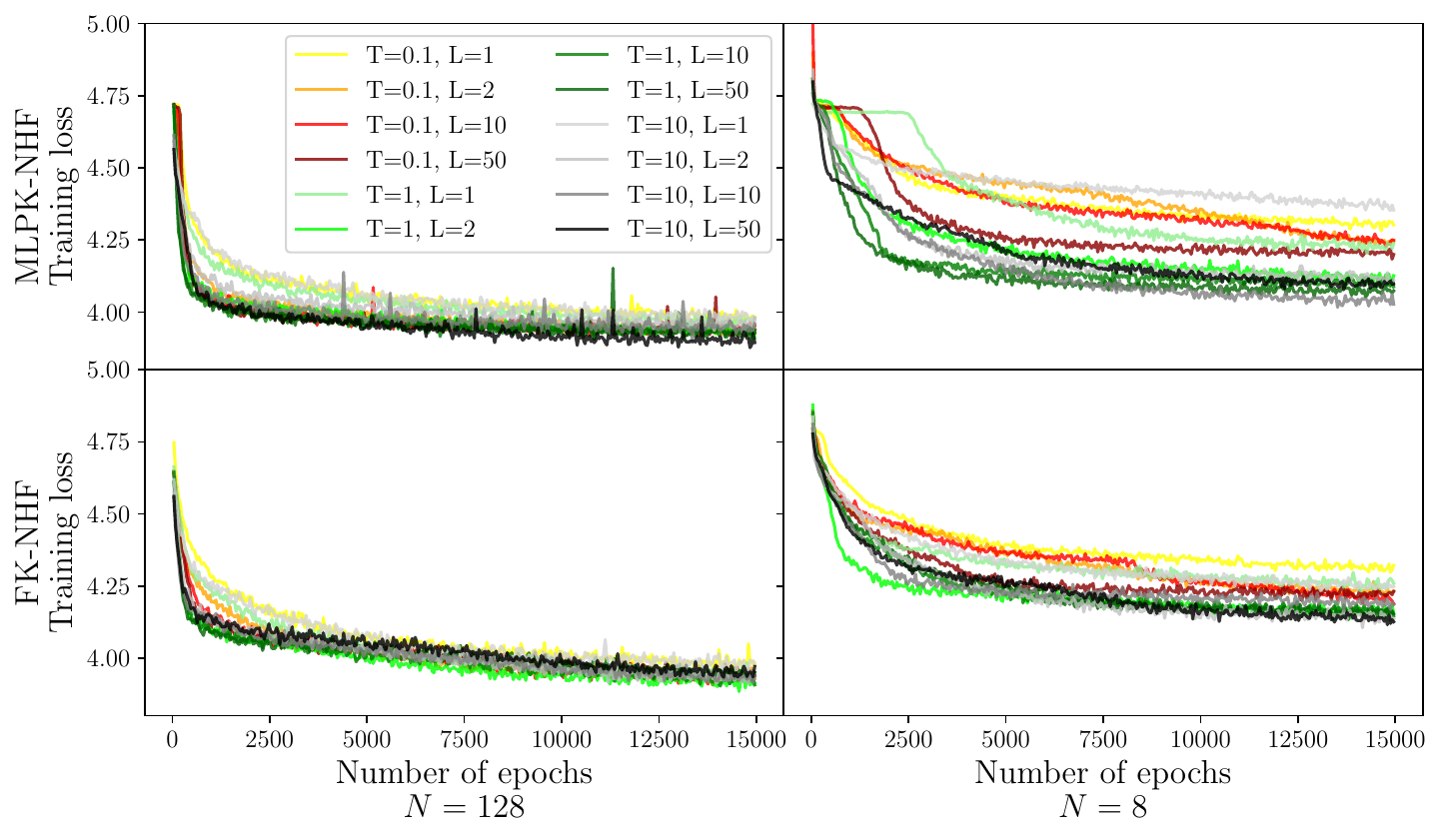}
\caption{Training loss as a function of epochs for models with different $N$ (number of neurons per hidden layer in each neural network of the model), $L$ (number of Leapfrog steps) and $T$ (integration time).}
\label{fig:comparisonLossesHTL}
\end{figure}

\subsection{Impact of Leapfrog-hyperparameters and model complexity}

Let us discuss the effect of Leapfrog-hyperparameters $L$ (number of Leapfrog steps) and $T = L \times \delta t$ (integration time) on the optimization, but also the impact of the model complexity. The latter is governed by the total number of neurons in the model, this number being an increasing function of $N$, the number of neurons per hidden layer in each MLP of the model. If the model is complex enough, we expect to learn how to adjust to the number of Leapfrog steps and choice of integration time. If not, the model may have better performance by increasing the number of steps, i.e. increasing $L$. 

We tested both FK-NHF and MLPK-NHF with various choices of $L$, $T$ and $N$ and a soft-uniform prior $\propto~s(x+3)s(-x+3)$, where $s$ is the sigmoid function. The corresponding loss decays are illustrated in Figure \ref{fig:comparisonLossesHTL} and additional details can be found in Appendix~\ref{subsec:addRobustness}.

First, FK-NHF is more robust than the MLPK one to the choices of $L$ and $T$, at fixed $N$, as discrepancy in the loss decay more clearly appears especially with $N=8$.
Then, regarding the tuning of the Leapfrog scheme, at fixed-integration time, models with $L=1$ always reach higher final value of the loss, this effect being less visible with FK-NHF. Increasing the number of leapfrog steps leads to better final performance even if the effect disappears once the number of Leapfrog steps gets sufficient and no further expressivity can be achieved. Finally, as for the effect of integration time $T$, it barely appears for FK-NHF, showing that the latter efficiently adjusts to this parameter. As for the MLPK-NHF, the effect of the integration time is clearer, but mostly at $N=8$, where performance improves for $T=1,10$ compared to $T=0.1$. Overall, as the number of parameters in the model increases, the impact of the integration time becomes limited. 

Thus, there are four hyperparameters that require tuning: three are usual in learning (minibatch size, learning rate and number of neurons per hidden layer, i.e. number of learning parameters of the model) and only one is specific to NHF: the number of Leapfrog steps, whose tuning is less sensitive when using FK-NHF. Furthermore, compared to diffusion models \citep{diffusionNetworks}, the required amount of steps is quite low.

\subsection{Impact of the prior distribution on the learned dynamics}

We illustrate the impact of the prior choice on the transfer of characteristics of the target distribution on the potential $V$, especially regarding the multimodality nature. All models were trained for 15,000 epochs using $N=128$, $T=1$ and $L=10$, with a 5,000 points training dataset and with the soft-uniform prior, a peaker Gaussian prior $\mathcal{N}(0,I_2)$  and a wide Gaussian prior $\mathcal{N}(0,2.5^2I_2)$. 

\begin{figure}[!h]
\centering
\includegraphics[width=1.0\hsize]{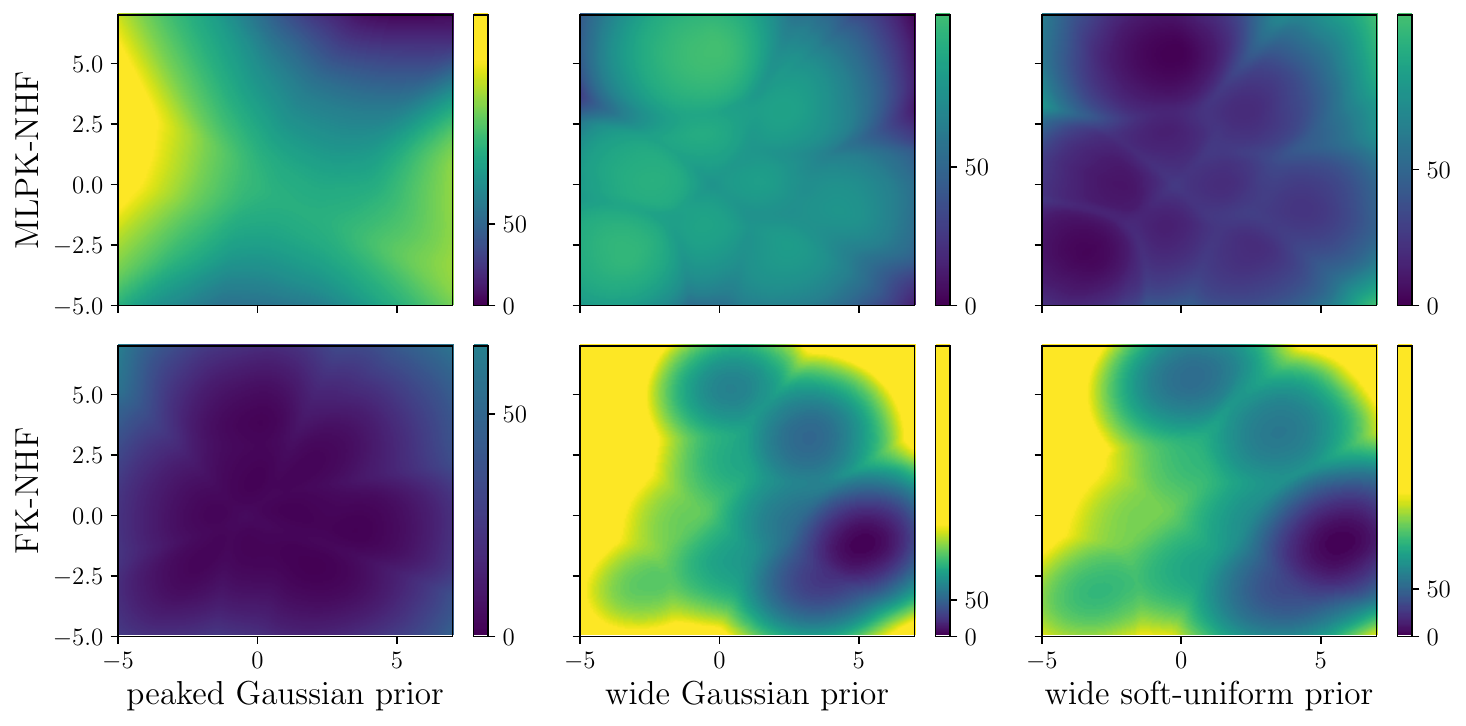}
\caption{Shifted potential energies learned by the six models previously defined.}
\label{fig:comparisonVshifted}
\end{figure}

All considered schemes recover the nine correct modes from the target distribution, as illustrated in Figure~\ref{fig:target}. We now consider the learned potential $V$. As the Hamiltonian evolution only involves its derivative, we represented a shifted version in Figure~\ref{fig:comparisonVshifted}. Choosing a relatively flat soft-uniform prior distribution that covers the target region, multimodality transfers to the potential energy for both FK-NHF and MLPK-NHF. The potential exhibits local extrema centered at the modes of the target, which can either be minima or maxima for the MLPK-NHF but are minima for the FK one. Indeed, with a MLPK model, the orientation of the learned energies may change from one numerical experiment to another, as we do not enforce the positiveness of the output of $V$ and $K$. Similar results were obtained using the wide  Gaussian prior with a variance large enough to cover the support of the target distribution, which stresses the impact of the spatial expansion rather than the nature of the prior distribution.

\begin{figure}[!h]
\centering
 \includegraphics[width=1.0\hsize]{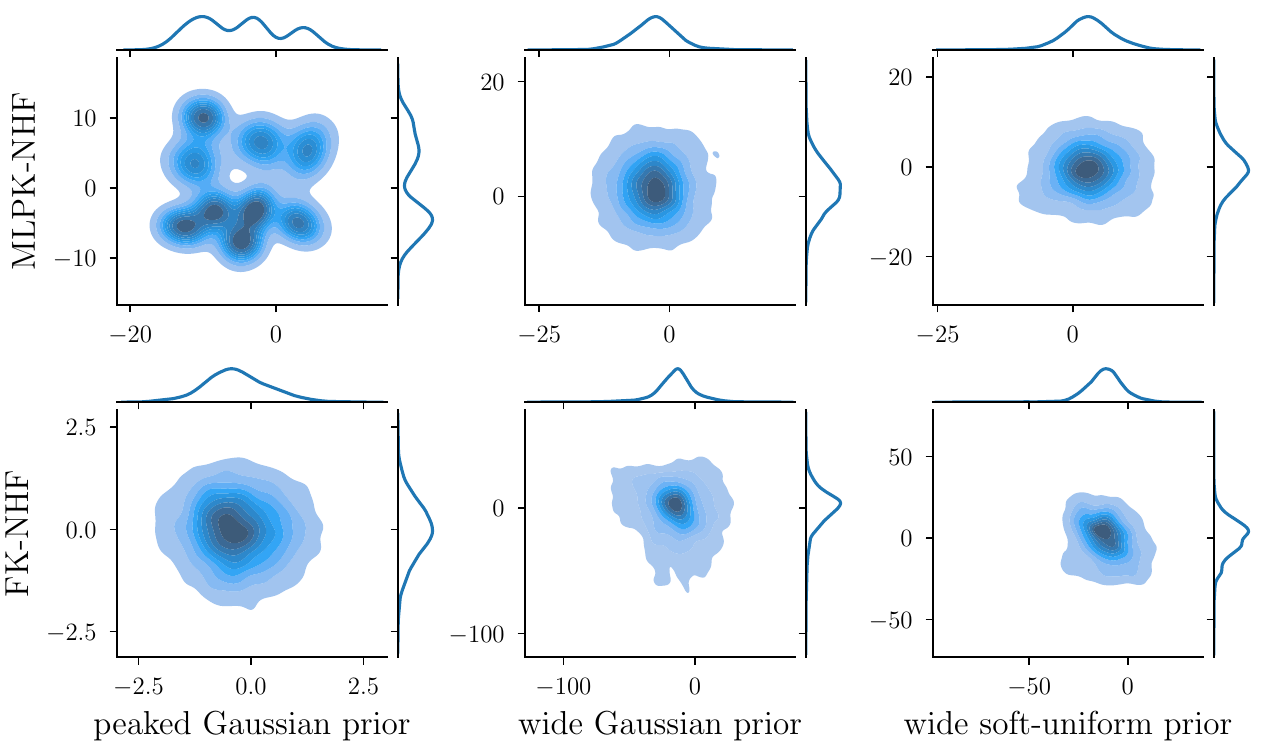} 
\caption{Density estimation of the artificial momenta from Encoder for the six models previously defined.}
\label{fig:comparisonpT}
\end{figure}

On the other hand, with a "peaked" prior distribution $\mathcal{N}(0,I_2)$ for the MLPK-NHF, the momenta $\mathbf{p}_T$ generated by the Encoder inherit from the multimodality of the target distribution, with the same number of modes (see Figure~\ref{fig:comparisonpT}). The learned energies are then different from the classical Physical ones and differ from one model to another. In the case of FK-NHF, multimodality is transferred to the potential energy, showing the robustness of the model to the choice of prior distribution.

Thus, using FK-NHF allows for a more robust transfer of important properties of the target distribution into the learned potential. More specifically, the model is able to learn an interpretable potential with extrema centered at the modes of the data. When the learned potential is not multimodal, it is an indication that multimodality has been transferred instead to the artificial momenta $\mathbf{p}_T$ generated by the Encoder.

\begin{figure}[!h]
    \centering
    \includegraphics[width=1.0\hsize]{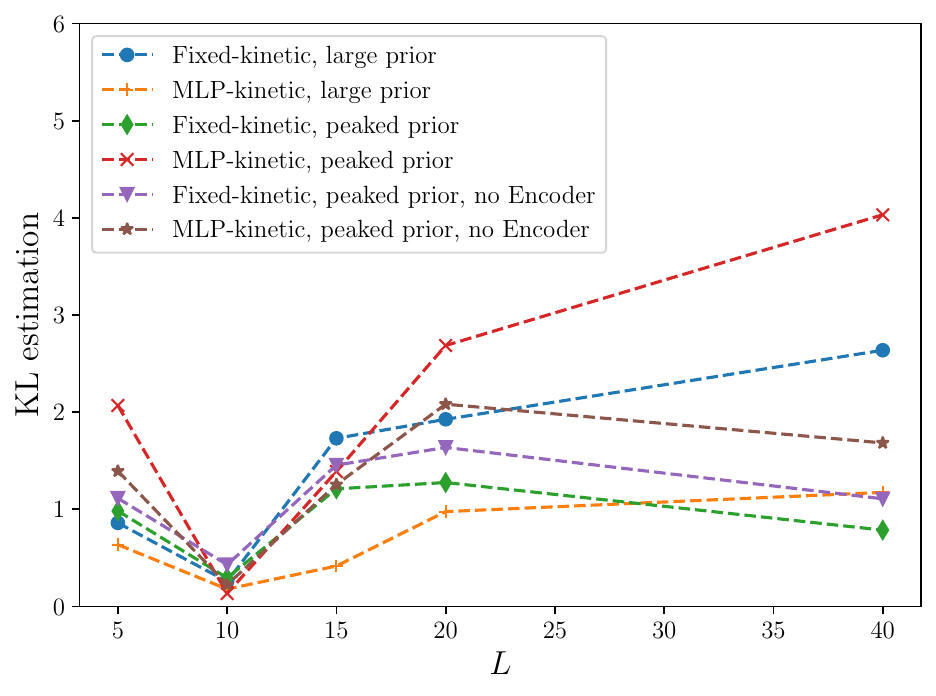}
    \caption{Evolution of KL estimation between the true and the model distribution as $L$ increases, for models trained with $L=10$. MLPK-NHF models with Encoder have $\sim 70,000$ learnable parameters while all the others have $\sim 50,000$ learnable parameters.}
\label{fig:kl_as_L_increases}
\end{figure}

\begin{figure}[!h]
    \centering
    \includegraphics[width=1.0\hsize]{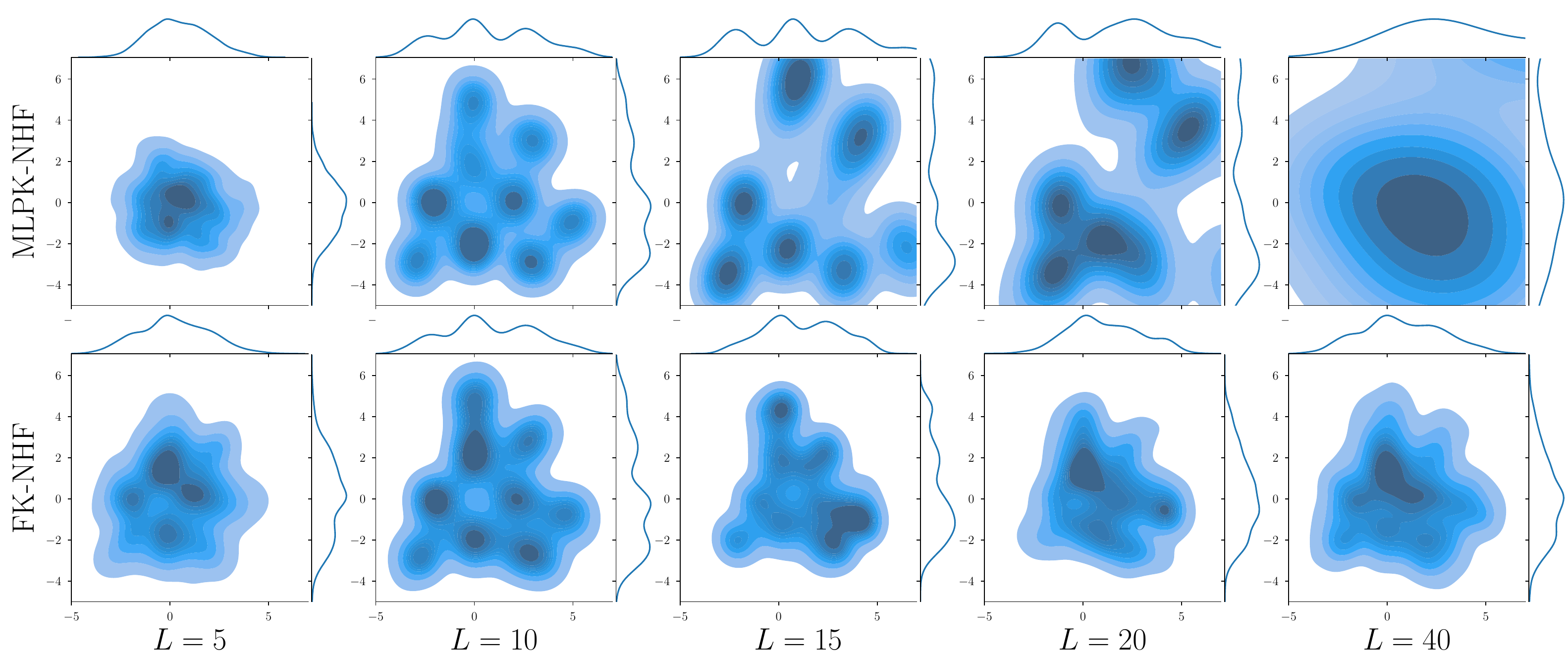}
    \caption{Density estimations as $L$ increases of samples generated by two models trained with $L = 10$ and a peaked Gaussian prior $\mathcal{N}(0,I_2)$. First row: MLPK-NHF which has learned a non-interpretable unimodal
potential. Second row: FK-NHF which has learned an interpretable multimodal potential.}
\label{fig:sampling_as_L_increases}
\end{figure}

Finally, learning a potential approximating $-\log \pi$ is interesting in terms of interpretability but also renders the model more robust to hyperparameters. Figures~\ref{fig:kl_as_L_increases} and \ref{fig:sampling_as_L_increases} illustrate how learning an interpretable multimodal potential makes the model less sensitive to the choice of Leapfrog steps and robust to some dynamics extrapolation.

As shown in Figure~\ref{fig:kl_as_L_increases} and Appendix~\ref{subsec:EncoderFree}, we also investigated the possibility of enforcing the transfer of multimodality to $V$ by removing the Encoder and having $\mathbf{p}_T$ drawn from a $\mathcal{N}(0,s^2I)$, $s$ being learned during training. While it improves MLPK-NHF for a peaked prior, we find it is less efficient and directly interpretable than fixing $K$. When fixing $K$, an Encoder-free model comes with a reduced complexity but we find that leaving as much flexibility as possible in the generation of momenta is the relevant option, especially for challenging problems and a small number of leapfrog steps.

\section{NHF FOR IMAGE GENERATION}\label{sec:HDresults}

\begin{figure}
    \centering
    \includegraphics[width=1.0\hsize]{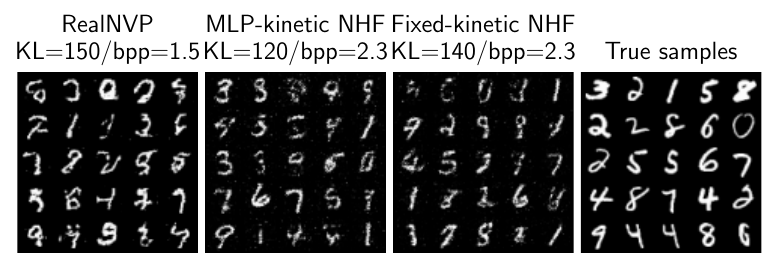}
    \includegraphics[width=1.0\hsize]{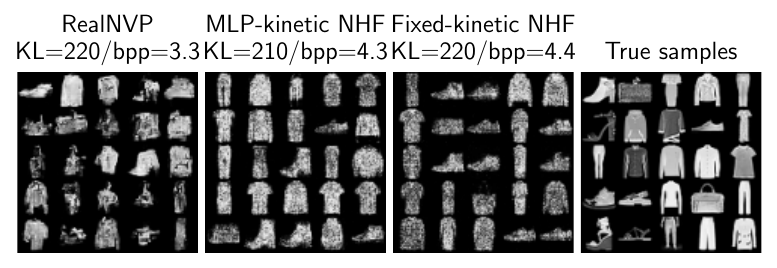}
    \caption{Samples produced after training on the MNIST and Fashion-MNIST datasets along with an estimation of the KL divergence between the true and model distributions and the number of bits per pixel.}
    \label{fig:mnist}
\end{figure}

To the best of our knowledge, NHF models have not been tested on
high-dimensional image generation problems. We run experiments for
sampling the MNIST handwritten digits \citep{digitsMNIST} dataset, as
well as additional tests on the Fashion MNIST dataset
\citep{fashionMNIST}. FK-NHF and MLPK-NHF are compared to a RealNVP
\citep{RealNVP} with a similar number of learnable parameters. The
RealNVP implementeation is based on an open-source GitHub repository
\footnote{\url{https://github.com/bjlkeng/sandbox/tree/master/realnvp}}. Details
on the experiments can be found in the
Appendix~\ref{subsec:expDetailsMNIST}. We use a pre-processing step
prescribed in \cite{RealNVP} consisting of learning the dequantized
target distribution in logit space. The choice of mass matrix for
FK-NHF is important for expressivity. This behaviour should be
familiar to the HMC community \citep{DUANE1987216} for which an
optimal mass matrix is important for efficient exploration
\citep{MassMatrixHMC}.

Quality of sampling is quantitatively assessed by the KL divergence,
estimated following \cite{estimationKL}, and the number of bits per
pixel \citep{bitsperdim}. Our experiments (Figure~\ref{fig:mnist} and
Appendix~\ref{subsec:hist_bit}) show that RealNVP and NHF slightly
outperform each other depending on the metric. The performance of
FK-NHF is achieved with a simpler architecture though.
Furthermore, the learned potentials have extrema located at the modes
of the target distribution, see
Appendix~\ref{subsec:InterpretabilityMNIST}. This capacity of the
models to learn the modes of the data and to store the information in
the potential energy underlines interpretability.

\begin{figure}[!h]
  \centering
  \includegraphics[width=0.49\hsize]{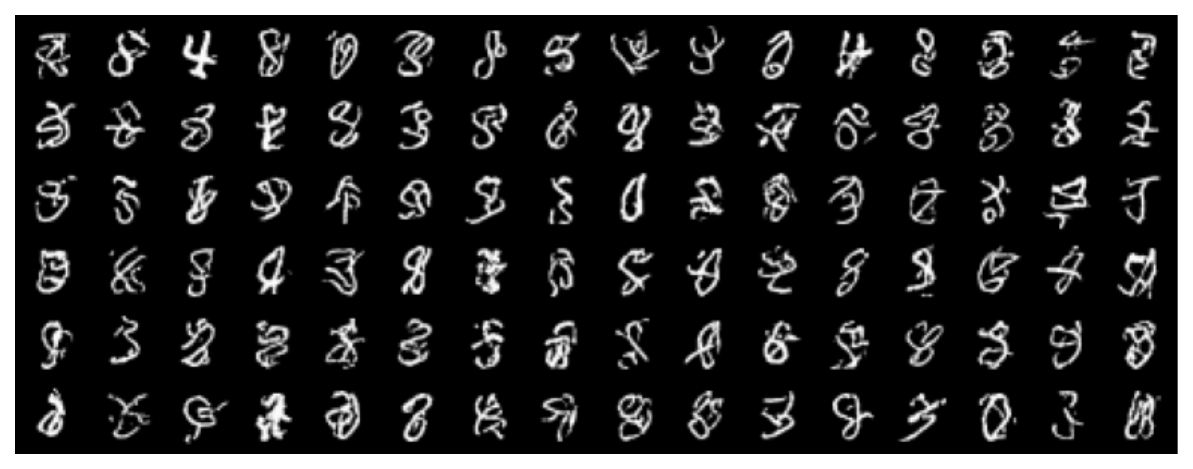}\hfill
  \includegraphics[width=0.49\hsize]{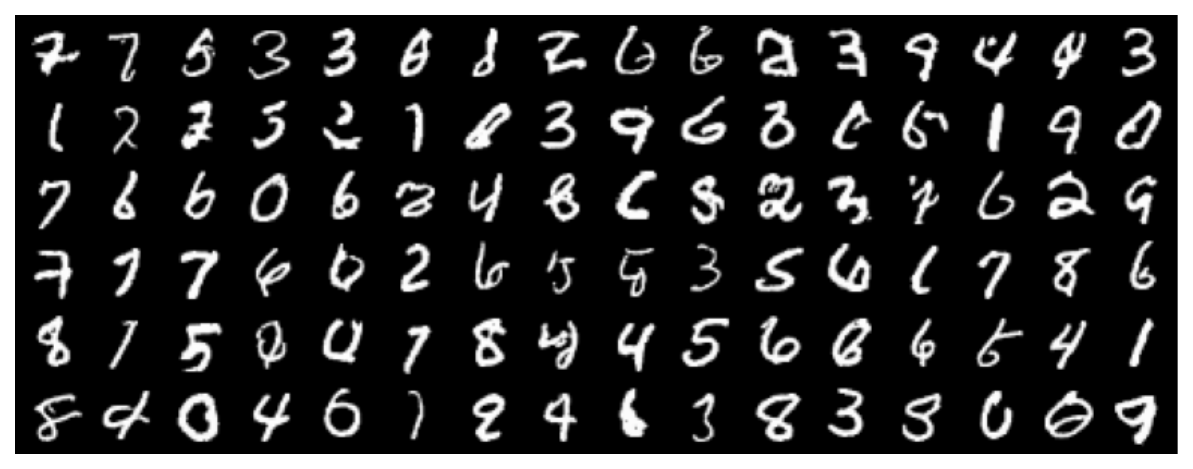}
  \includegraphics[width=0.49\hsize]{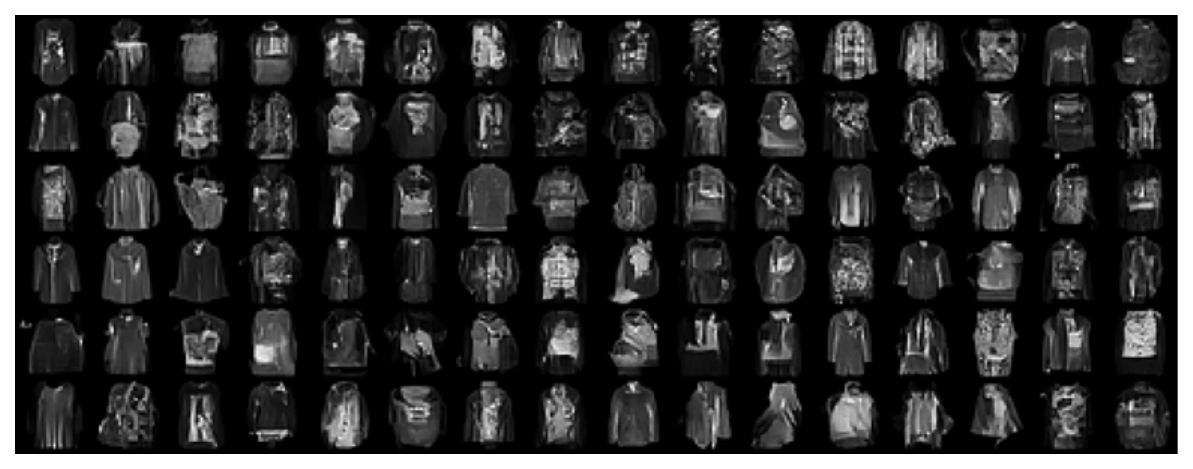}
\hfill
  \includegraphics[width=0.49\hsize]{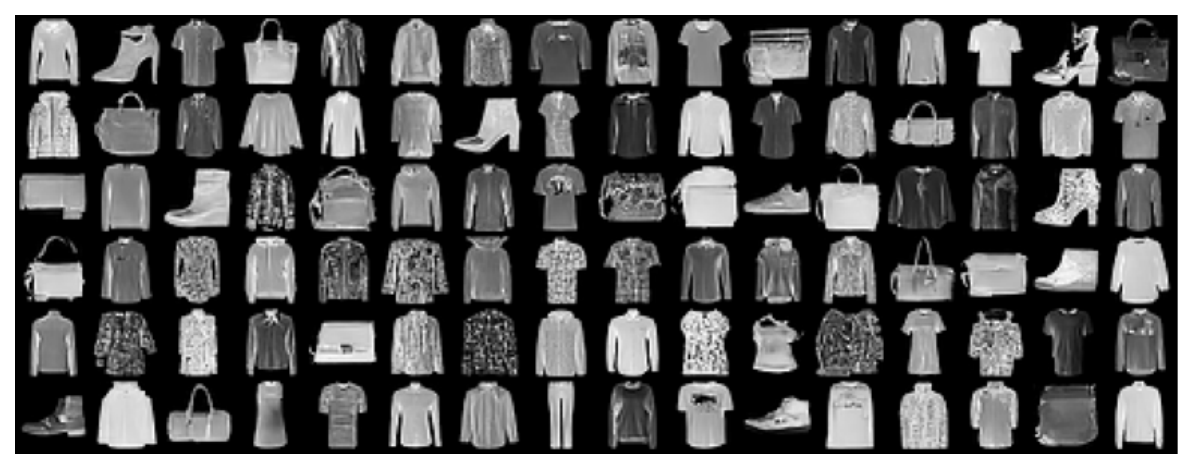}
\caption{Samples produced by the diffusion model after training on the MNIST and Fashion-MNIST datasets. Left: $L=10$ denoising steps (KL divergence between the true and model distributions: KL$=250$ (MNIST), KL=$340$ (Fashion-MNIST)). Right: $L=100$ denoising step (KL$=70$ (MNIST), KL$=70$ (Fashion-MNIST)).}
\label{fig:diff}
\end{figure}

To compare ODE-driven flow-based models, such as NHF, with SDE-driven
diffusion models \citep{diffusionNetworks, ho}, we also tested an
implementation of \cite{ho} based on an open-source GitHub repository
\footnote{\url{https://github.com/lucidrains/denoising-diffusion-pytorch/}}.
We adapted it so that it has 2.7 million parameters (same order as the
other algorithms) and uses the same pre-processing step. The
experiments ran on a HPC cluster with the same number of epochs and
batch size than for the previous experiments with normalizing
flows. We tested the model with $L=10$ and $L=100$ denoising
steps. Results are visible on Figure~\ref{fig:diff} and it appears
that around $100$ denoising steps are required for producing good
samples. This should be directly compared with the 10 Leapfrog steps
used within NHF models. This comes in addition to the extra cost of
sampling with a trained diffusion model since it requires
approximating the reverse process from noise to images with a Markov
chain.

\section{ADAPTING NHF FOR BAYESIAN INFERENCE}
\label{sec:HNFforBayes}

\subsection{Methodology, derivation of the new loss function}

\begin{figure}[!h]
    \centering
    \includegraphics[width=1.0\hsize]{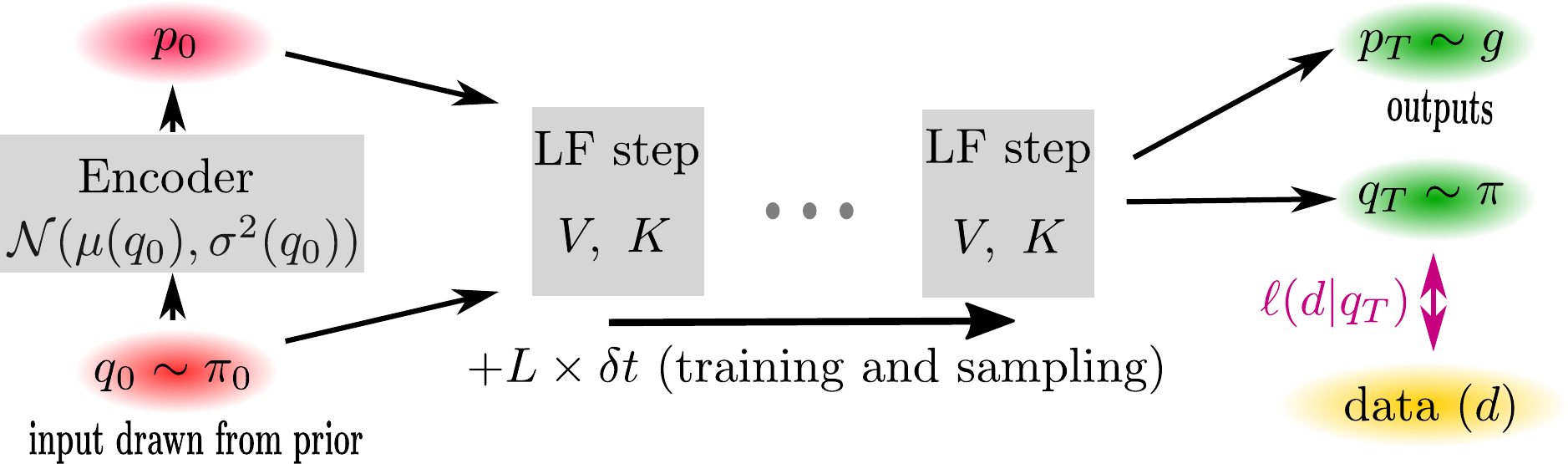}
    \caption{Schematic representation of NHF for Bayesian inference.}
    \label{fig:inferenceNHF}
\end{figure}

NHF can be used to perform Bayesian inference, by using Hamiltonian
flows to transform the prior distribution, in the sense of Bayes'
theorem, $\pi_0$ of some vector of parameters $\mathbf{q}$ into the
target posterior distribution $\pi(\mathbf{q}|\boldsymbol{d})$ of
these parameters, knowing some data $\boldsymbol{d}$ and likelihood
distribution $\ell$ (see Figure~\ref{fig:inferenceNHF}). The main
difference with the above-described NHF lies in the loss inspired by
the KL phase in Boltzmann Generators \citep{boltzmannGenerators}, as
well as in the learning procedure. The NHF becomes a generator of a
family of functions for variational inference. During training, this
NHF takes batches of $\mathbf{q}_0$ from the prior distribution as
inputs. For each $\mathbf{q}_0$, one $\mathbf{p}_0$ is drawn from a
Gaussian distribution whose mean and deviation depend on the
$\mathbf{q}_0$. The resulting point in phase-space evolves through $L$
Leapfrog steps with integration time $\delta t$. The outputs consist
in the final positions $\mathbf{q}_T$ and momenta $\mathbf{p}_T$, as
well as the initial mean $\mu(\mathbf{q}_0)$, deviation
$\sigma(\mathbf{q}_0)$ and $\mathbf{p}_0$. All these outputs, as well
as the data $\boldsymbol{d}$, are used in the loss computation. Once
trained, it can transform the prior into the desired posterior
distribution of the parameters. Thus, both training and sampling are
now made following the forward-direction flow from the prior to the
posterior.

Computing the loss requires access to the likelihood distribution $\ell$ of the model, which encapsulates the covariance matrix of the data as well as the underlying physical mapping between vectors of parameters and the corresponding data. In the framework of Hamiltonian dynamics, the full system is made of both positions (the parameters of interest) and artificial momenta. We call $\mathbf{q}_0, \mathbf{p}_0$ the initial position and momentum, respectively, and $\mathbf{q}_T, \mathbf{p}_T$ the corresponding final position and momentum, respectively, obtained after $L$ Leapfrog transformations $\mathcal{T}^{\delta t}_1,...,\ \mathcal{T}^{\delta t}_L$ with timestep $\delta t$, i.e: $(\mathbf{q}_T,\mathbf{p}_T) = \mathcal{T}^{\delta t}_L \circ ... \circ \mathcal{T}^{\delta t}_1 (\mathbf{q}_0,\mathbf{p}_0) :=  \mathcal{T}(\mathbf{q}_0,\mathbf{p}_0)$. Also, we introduce the notations for the projections along the final positions and momenta, i.e. $\mathbf{q}_T := \mathcal{T}_q(\mathbf{q}_0,\mathbf{p}_0)$ and $\mathbf{p}_T := \mathcal{T}_p(\mathbf{q}_0,\mathbf{p}_0)$. By changing the variables, 
the model joint distribution $M$ may be written as:
\begin{align*}
    M(\mathbf{q}_T,\mathbf{p}_T) &= 1 \times \Pi_0(\mathcal{T}^{-\delta t}_1 \circ ... \circ \mathcal{T}^{-\delta t}_L (\mathbf{q}_T,\mathbf{p}_T)) \\
    &= \Pi_0(\mathbf{q}_0,\mathbf{p}_0) = \pi_0(\mathbf{q}_0) \times f(\mathbf{p}_0|\mathbf{q}_0),
\end{align*}
where $\Pi_0$ is the joint prior distribution, $\pi_0$ the prior distribution of the parameters of interest and $f$ the Gaussian distribution of the Encoder. We fix the target density of the final momenta $g(\mathbf{p})$ (e.g. Gaussian). We then minimize the KL-divergence between the model joint distribution and the desired target joint distribution conditioned on data $\Pi(\mathbf{q},\mathbf{p} | \boldsymbol{d}) = \pi(\mathbf{q}|\boldsymbol{d}) g(\mathbf{p})$. We write the latter as the product of a density depending on $\mathbf{q}$ and one depending on $\mathbf{p}$. Using Bayes' theorem, we have (see Appendix~\ref{app:lossesInference}):
\begin{equation}\label{KL}
\begin{split}
    &D_{KL}(M(\mathbf{q}_T,\mathbf{p}_T) \ || \ \pi(\mathbf{q}_T|\boldsymbol{d}) g(\mathbf{p}_T))=\int \pi_0(\mathbf{q}_0,\mathbf{p}_0) \\
    & \times\big[\log \pi_0(\mathbf{q}_0) \!+\!\log f(\mathbf{p}_0|\mathbf{q}_0) \!-\! \log \pi_0(\mathcal{T}_q(\mathbf{q}_0,\mathbf{p}_0)) \\
    & \!-\! \log \ell(\boldsymbol{d} | \mathcal{T}_q(\mathbf{q}_0,\mathbf{p}_0)) \!-\! \log g(\mathcal{T}_p(\mathbf{q}_0,\mathbf{p}_0)\big] d\mathbf{q}_0d\mathbf{p}_0 \!+\! \text{cst}.
\end{split}
\end{equation}

\subsection{Application to cosmology}

We apply the above architecture to cosmological analysis: the determination of the cosmic expansion, and more generally of the cosmological parameters, from the observation of brightness and recession velocity of Type Ia supernov\ae{} \cite[e.g.][]{Riess1998,Betoule_2014}. While this model used so far has been simple, it may be expanded in very complicated directions for which sampling from the probability distribution becomes complex. New observatories are presently being built and expected to deliver tens of thousands of new supernov\ae{} Ia over the next decade \citep{LSST}.

According to the $\Lambda$-CDM model, the relation between the distance and the brightness of Type Ia supernovae is of great interest because it depends on two cosmological parameters: the matter density parameter $\Omega_m$ and the adimensional Hubble parameter $h$. To be more specific, a database of type Ia supernovae reports the distance modulus $\mu$. This quantity is defined as the difference between the apparent and the absolute magnitude of an astronomical object and is directly related to luminosity distance \citep{Weinberg1972} and thus a function of the redshift $z$, $\Omega_m$ and $h$:
$$	\mu(z,\Omega_m,h) = 5\log_{10}\left(\frac{D_L^*(z,\Omega_m)}{h\; 10\text{pc}}\right) 
$$
where $D_L^*$ is a function for which one can get a closed-form approximation in a flat Universe \citep{Pen_1999}, see Appendix~\ref{app:CosmoDetails} for details. 

\begin{figure}[!h]
\centering
 \includegraphics[width=1.0\hsize]{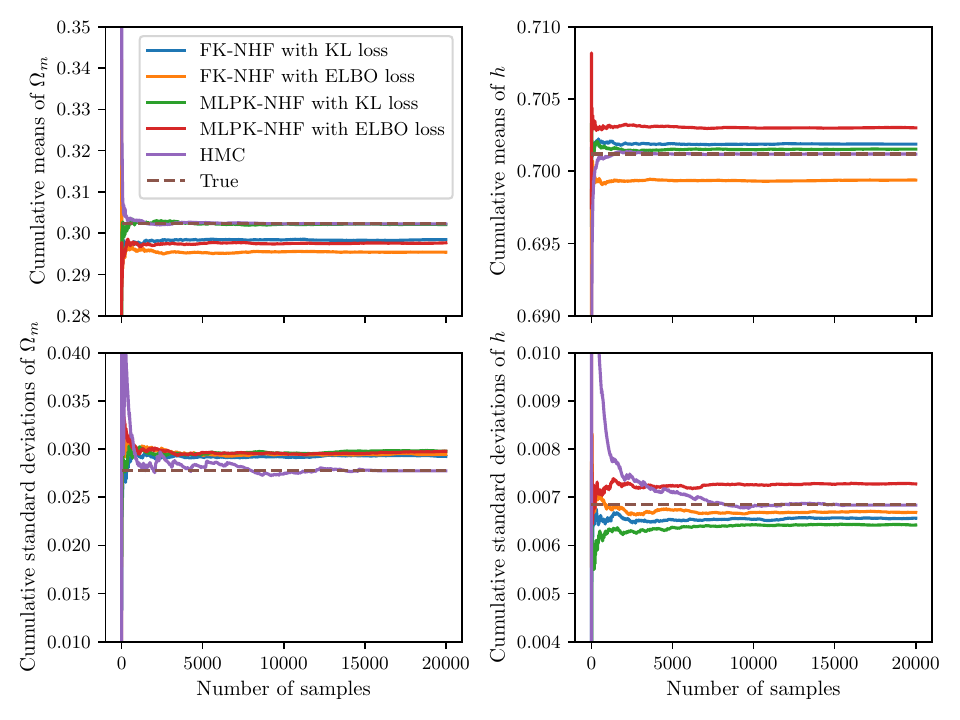} 
\caption{Trace plots of means and standard deviations of $\Omega_m$ and $h$ produced by trained NHF models and for an HMC on a 20,000-sample dataset, compared to the ground truth. Soft-uniform prior, 30,000 training epochs, $g\sim\mathcal{N}(0, I_2)$.}
\label{fig:cumulativePlotsOh}
\end{figure}

We aim to sample from the posterior distribution $\pi(\Omega_m,h|\text{data})$ quantifying the probability that we are living in a universe whose mean density and expansion is equal to $\Omega_m$ and $h$ given $D$ observations $\text{data}=\{z_i, \mu_i\}_{1 \leq i \leq D}$ of type Ia supernovae, and the covariance matrix $C$ of the observed distance moduli. We assume for simplicity that the likelihood of the problem is Gaussian, i.e. the observed data and the simulated output from parameters differ up to Gaussian noise. We note that the exact simulation for cosmological analysis of the supernova brightness is a complicated and expensive procedure, involving many nuisance parameters which participate in the final noise.  It also acts as an example for more complex inference procedures, such as one relying on galaxy clustering, or weak lensing. It is thus crucial to use the least amount of parameters, and the least simulations possible to run the inference, which is the aim of this section. The final momenta distribution $g$ is set to a Normal distribution. The trace plots in figure~\ref{fig:cumulativePlotsOh} compare the performance of an FK and MLPK-NHF with an HMC. They represent the cumulative means and standard deviations of the set of samples. These experiments first show that NHF, using learnt Hamiltonian flows, are competitive with HMC, using exact Hamiltonian flows. Indeed, NHF samples only show around $1\%$ bias error but they are completely uncorrelated from each other. After training, for a fixed-computational budget, the convergence of the empirical average can then be faster than with HMC (clearly appearing for instance on the variance of $h$ in figure~\ref{fig:cumulativePlotsOh}). Furthermore, the experiments show that the KL-minimization approach is performing better. We leave for future work a possible correction using importance sampling methods at the end of training. More generally, a careful analysis of the trade-off between sampling quality and computational cost, typically on a complex and multimodal target, will be of interest.

\section{CONCLUSION}

In this work, we analyzed and improved Normalizing Hamiltonian Flows algorithms for Generative modeling. The main advantage of these methods is twofold. First, the volume-preservation in phase-space avoids the costly computation of Jacobian determinants. Then, as reversibility is ensured by the symplectic integrator, they allow for flexibility in the neural network architecture. This flexibility allowed us to propose a NHF variant based on classical kinetic energy. By exploring a 2D mixture problem, we illustrated how the explicit classical design of the kinetic energy is a way to increase robustness and facilitate interpretability while reducing the computational cost. While testing NHF models for image generation, both show similar generative performance and are able to preserve their interpretability properties. It is noteworthy that, compared to diffusion models, they only require a short dynamics integration. Finally, we explained how to adapt NHF to the context of Bayesian inference to obtain a sampler of the posterior distribution. Further work will address methodological issues as to how the bias generated by a trained model could be corrected by importance sampling techniques, typically on high dimensional cosmological models but also more fundamental questions regarding a more precise comparison of NHF with diffusion models.

\section*{Acknowledgments}
All the authors are grateful for the support from CNRS through the MITI-Prime 80 project "CosmoBayes". All the authors thank the Mésocentre Clermont Auvergne University (\url{https://mesocentre.uca.fr/}), and the HPC cluster at Fysikum (\url{it.fysik.su.se/hpc/}) for providing help, computing and storage resources. This work was performed using HPC resources from GENCI–IDRIS (Grant 2023-AD011014013). It was supported by the French ANR under the grant ANR-20-CE46-0007 (\textit{SuSa} project), by the Simons Collaboration on ``Learning the Universe'' and by Institut Pascal at Université Paris-Saclay with the support of the program Investissements d’avenir ANR-11-IDEX-0003-01. AG acknowledges the support of the Institut Universitaire de France. JJ acknowledges support from the Swedish Research Council (VR) under the project 2020-05143 -- ``Deciphering the Dynamics of Cosmic Structure". This work was carried out within the Aquila consortium (\url{https://www.aquila-consortium.org}).

\bibliography{biblio2}
\bibliographystyle{apalike}

\newpage
~\newpage

\textbf{\Large{Checklist}}

 \begin{enumerate}

 \item For all models and algorithms presented, check if you include:
 \begin{enumerate}
   \item A clear description of the mathematical setting, assumptions, algorithm, and/or model. [\textbf{Yes}/No/Not Applicable] 
   \item An analysis of the properties and complexity (time, space, sample size) of any algorithm. [\textbf{Yes}/No/Not Applicable]
   \item (Optional) Anonymized source code, with specification of all dependencies, including external libraries. [\textbf{Yes}/No/Not Applicable] \textbf{We attached a ready-to-run Python script for testing a NHF during submission. Also, a public version of the code \footnote{\url{https://plmlab.math.cnrs.fr/stoch-algo-phys/generative-models/fixed-kinetic-NHF/}} has been released.}
 \end{enumerate}

 \item For any theoretical claim, check if you include:
 \begin{enumerate}
   \item Statements of the full set of assumptions of all theoretical results. [\textbf{Yes}/No/Not Applicable]
   \item Complete proofs of all theoretical results. [\textbf{Yes}/No/Not Applicable] \textbf{Mathematical derivations can be found in the Appendices.}
   \item Clear explanations of any assumptions. [\textbf{Yes}/No/Not Applicable]     
 \end{enumerate}

 \item For all figures and tables that present empirical results, check if you include:
 \begin{enumerate}
   \item The code, data, and instructions needed to reproduce the main experimental results (either in the supplemental material or as a URL). [\textbf{Yes}/No/Not Applicable] \textbf{See Appendices.}
   \item All the training details (e.g., data splits, hyperparameters, how they were chosen). [\textbf{Yes}/No/Not Applicable] \textbf{See Appendices.}
         \item A clear definition of the specific measure or statistics and error bars (e.g., with respect to the random seed after running experiments multiple times). [\textbf{Yes}/No/Not Applicable] \textbf{See Appendices.}
         \item A description of the computing infrastructure used. (e.g., type of GPUs, internal cluster, or cloud provider). [\textbf{Yes}/No/Not Applicable] \textbf{See Appendices.}
 \end{enumerate}

 \item If you are using existing assets (e.g., code, data, models) or curating/releasing new assets, check if you include:
 \begin{enumerate}
   \item Citations of the creator If your work uses existing assets. [\textbf{Yes}/No/Not Applicable]
   \item The license information of the assets, if applicable. [\textbf{Yes}/No/Not Applicable]
   \item New assets either in the supplemental material or as a URL, if applicable. [\textbf{Yes}/No/Not Applicable]
   \item Information about consent from data providers/curators. [\textbf{Yes}/No/Not Applicable]
   \item Discussion of sensible content if applicable, e.g., personally identifiable information or offensive content. [Yes/No/\textbf{Not Applicable}]
 \end{enumerate}

 \item If you used crowdsourcing or conducted research with human subjects, check if you include:
 \begin{enumerate}
   \item The full text of instructions given to participants and screenshots. [Yes/No/\textbf{Not Applicable}]
   \item Descriptions of potential participant risks, with links to Institutional Review Board (IRB) approvals if applicable. [Yes/No/\textbf{Not Applicable}]
   \item The estimated hourly wage paid to participants and the total amount spent on participant compensation. [Yes/No/\textbf{Not Applicable}]
 \end{enumerate}

 \end{enumerate}

\onecolumn
\begin{center}
    \Large{\textbf{Supplementary Materials}}
\end{center}

\vspace{1\baselineskip}

\begin{appendix}

\section{METRICS FOR EVALUATING THE QUALITY OF SAMPLING}\label{app:metrics}

In a $D$-dimensional space, we consider a set of $S$ samples $\{X_s\}_{s=1}^S := \{(X_{s,1},...,X_{s,D})\}_{s=1}^S$ generated by a model and $S$ samples $\{X_s^0\}_{s=1}^S := \{(X_{s,1}^0,...,X_{s,D}^0)\}_{s=1}^S$ drawn from the true (test) dataset.

\subsection{Kullback-Leibler divergence}\label{subsec:KL}

A natural (pseudo-)distance is the KL-divergence $D_{KL}(\pi||m)$ between the true target distribution with density $\pi$ and the model distribution with density $m$. This pseudo-distance quantifies the loss of information when using the model distribution instead of the true target for describing the data - so the lower the better. The KL-divergence is defined as
$$D_{KL}(\pi || m) = \int \pi(x) \ln \frac{\pi(x)}{m(x)} dx \geq 0.$$

When one of these probability distributions is intractable, as it is the case here, we estimate $D_{KL}$ by comparing samples from the true target distribution with samples from the model distribution. The procedure described in \cite{estimationKL} proceeds according to a $k$-neighborhood density estimate of the two distributions:
$$D_{KL}(\pi||m) \approx -\frac{D}{S} \sum_{s=1}^S \ln \frac{r_k(X_s)}{s_k(X_s)} + \ln \frac{S}{S-1}$$
where $r_k(X_s)$ is the $k$-th closest neighbor of $X_s$ in $\{X_s\}_{s=1}^S \backslash \{X_s\}$ and $s_k(X_s)$ is the $k$-th closest neighbor of $X_s$ in $\{X^0_s\}_{s=1}^S$. Parameters chosen in our estimation are $S=1024$ and $k=1$.

\subsection{Bits per pixel}\label{subsec:bpp}

For evaluating the quality of images generated by the considered models, we compute the number of bits per pixel, decaying as the quality of the image increases. We start with a pre-processing step. First, pixels of training images are dequantized by adding uniform noise $\varepsilon \sim \mathcal{U}]0,1[$ and ranging them back to interval $[0,1]$ as $x \leftarrow (255x + \varepsilon)/256$. Then, the models are trained on the target distribution in logit space by transforming the resulting noisy pixels as $x \leftarrow \text{logit}((1-2\lambda)x + \lambda)$ with $\lambda = 10^{-6}$. 

Once trained, we evaluate the number of bits per pixel of a pre-processed image $\Tilde{X}^0_s := (\Tilde{X}^0_{s,1},...,\Tilde{X}^0_{s,D})$ in logit space from the test dataset following \cite{bitsperdim}: 
$$b(\Tilde{X}^0_s) = - \frac{\ln m(\Tilde{X}^0_s)}{D \ln 2} - \log_2(1-2\lambda) + \frac{1}{D} \sum_{d=1}^D \left[ \log_2(\text{logit}(\Tilde{X}^0_{s,d})) + \log_2(1-\text{logit}(\Tilde{X}^0_{s,d})) \right].$$
In the above equation, we evaluate the probability distribution in the logit space of the model, namely $m(\Tilde{X}^0_s)$. For a classical flow-based model, it is straightforward since $m(\Tilde{X}^0_s) = \pi_0(T^{-1}(\Tilde{X}^0_s)) \times \left| \text{Jac}_{T^{-1}}(\Tilde{X}^0_s) \right|$ where $T^{-1}$ is the transformation from logit space to latent space learned by the model. 

However, for NHF, the change of variable formula is only valid in phase-space for the model joint distribution of the positions and momenta: $M(\Tilde{X}^0_s, V_s) = \Pi_0(T^{-1}(\Tilde{X}^0_s, V_s)) \times 1$. We use a Monte Carlo approximation of $m(\Tilde{X}^0_s)$ by drawing $N$ momenta $V_{s,1},...,V_{s,N}$ from the Gaussian distribution $f(.|\Tilde{X}^0_s)$ parameterized by the Encoder as:
$$m(\Tilde{X}^0_s) = \int M(\Tilde{X}^0_s,V_s)dV_s \approx \frac{1}{N} \sum_{i=1}^N \frac{M(\Tilde{X}^0_s,V_{s,i})}{f(V_{s,i}|\Tilde{X}^0_s)}.$$
For evaluating the number of bits per pixel of a model on the two MNIST datasets, we average the bits per pixel values obtained with $1024$ images from the test dataset, using $N=10$ for NHF. The histograms of bits per pixel are exhibited in Appendix~\ref{subsec:hist_bit}.

\section{NUMERICAL EXPERIMENTS ON THE 2D PROBLEM}
\label{app:experiments2D}

\subsection{Experimental details}\label{subsec:expDetails2D}

For the 2D Gaussian mixture problem, we tested four different NHF models:
\begin{itemize}
    \item MLP-kinetic NHF with Encoder: $\mu$ and $\sigma$ are MLPs with size $(2, N, N, 2)$, $V$ and $K$ are MLPs with size $(2, N, N, 1)$. According to the experiments, we used $N=8, \ 32, \ 128$.
    \item Fixed-kinetic NHF with Encoder: $\mu$ and $\sigma$ are MLPs with size $(2, N, N, 2)$, $V$ is a MLP with size $(2, N, N, 1)$. Kinetic energy $K$ is a positive quadratic form whose mass matrix is learned during training. According to the experiments, we used $N=8, \ 32, \ 128$.
    \item MLP-kinetic NHF without Encoder: we use this model for experiments in Section 5. Artificial momenta $\mathbf{p}_T$ are drawn from a $\mathcal{N}(0,C)$ where $C = \text{Diag}(s_1^2,s_2^2)$, $s_1, s_2$ being learned during training. $V$ and $K$ are MLPs with size $(2, 156, 156, 1)$. The number of neurons per hidden layer was chosen so that the resulting model has about the same number of parameters as an Encoder-based Fixed-kinetic NHF with $N=128$.
    \item Fixed-kinetic NHF without Encoder: this model is considered in Section 5. Artificial momenta $\mathbf{p}_T$ are drawn from a $\mathcal{N}(0,C)$ where $C = \text{Diag}(s_1^2,s_2^2)$, $s_1, s_2$ being learned during training. $V$ is a MLP with size $(2, 220, 220, 1)$. Kinetic energy $K$ is a positive quadratic form whose mass matrix is learned during training. The number of neurons per hidden layer was chosen so that the resulting model has about the same number of parameters as an Encoder-based Fixed-kinetic NHF with $N=128$.
\end{itemize}

We used Softplus activation functions in between hidden layers. All models were trained on a 5,000
points dataset with minibatches of size 512. Weights and biases were optimized with the Adam algorithm \cite{Adam}, setting the learning rate to $5 \times 10^{-4}$. The experiments were run on a HPC cluster, each of them using one GPU.

\subsection{Additional experiments for showing robustness of the models}\label{subsec:addRobustness}

\begin{figure}[!ht]
\centering
\includegraphics[width=0.8\hsize]{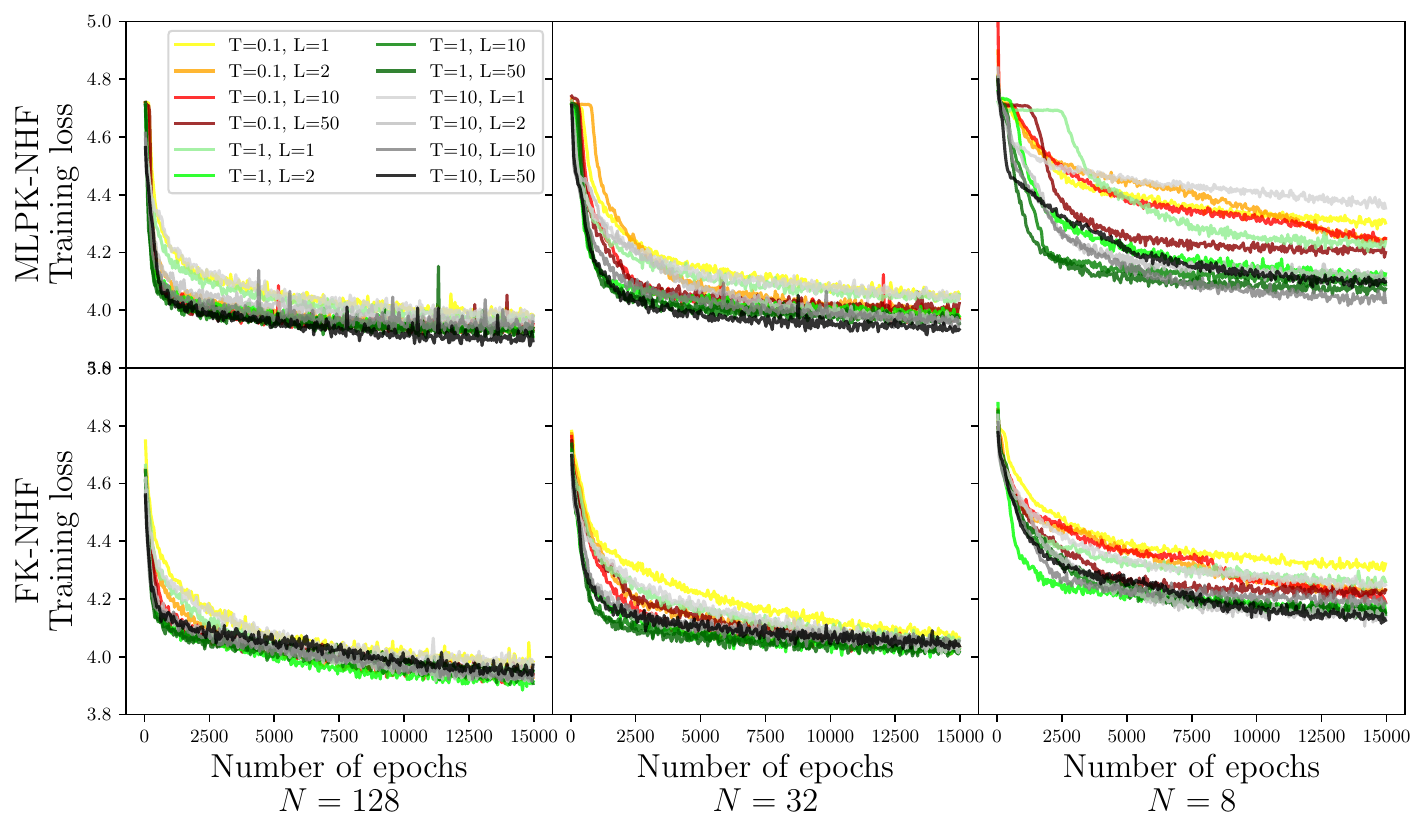}
\caption{Training loss as a function of epochs for models with different $N$ (number of neurons per hidden layer in each neural network of the model), $L$ (number of Leapfrog steps) and $T$ (integration time).}
\label{fig:comparisonLossesHTL2}
\end{figure}

\begin{figure}[!ht]
\centering
\includegraphics[width=0.6\hsize]{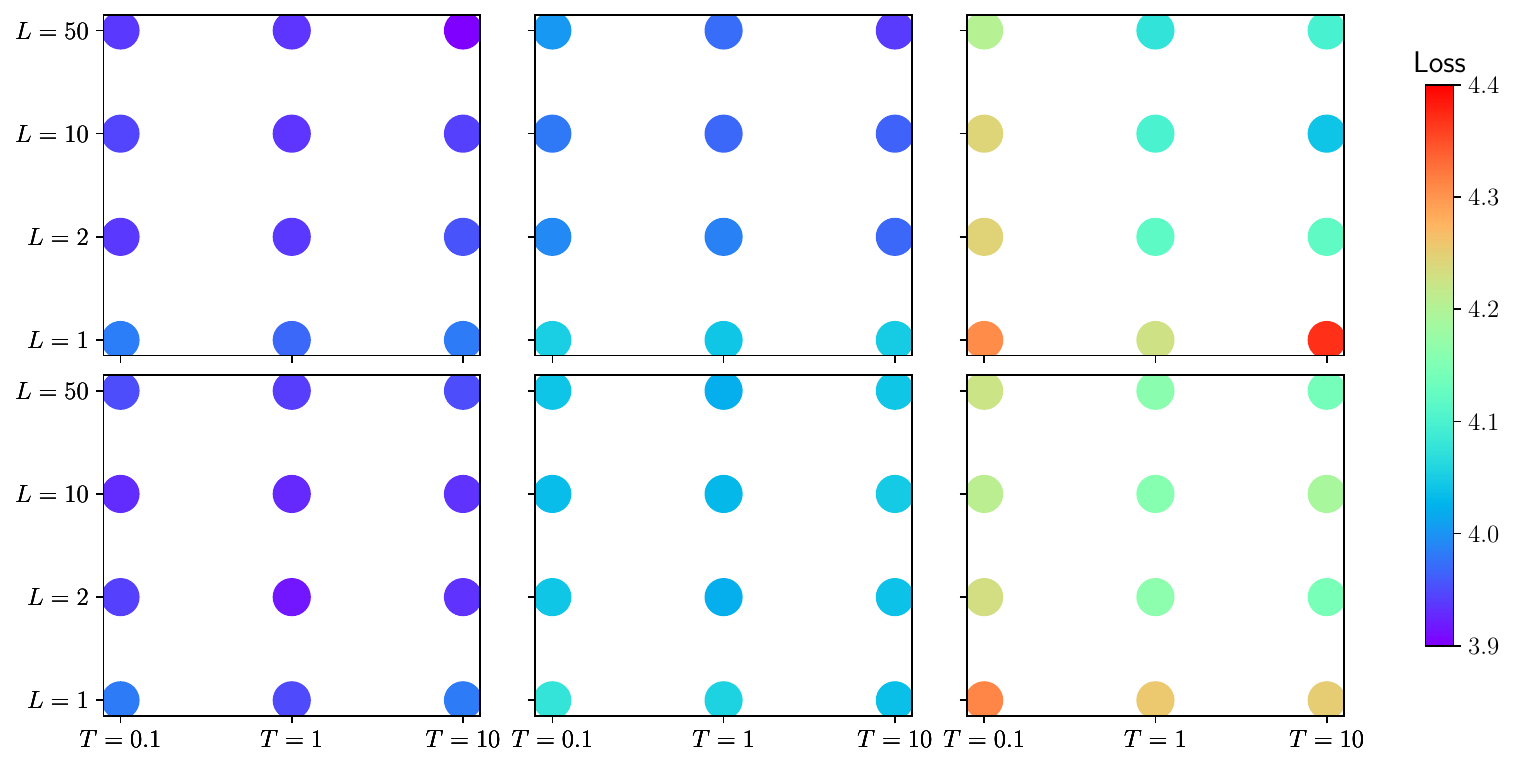}
\caption{Scatter plots $L$ vs. $T$ of final values of the loss averaged on last 500 epochs. First row: MLP-kinetic NHF; second row: fixed-kinetic model. From left to right: $N=128, \ 32, \ 8$. Models were trained for $15,000$ epochs on a 5,000-point dataset with minibatches of size $512$.}
\label{fig:comparisonLossesTL2}
\end{figure}

We also present the results of additional experiments with $N=32$ in Figure \ref{fig:comparisonLossesHTL2}. Removing more than 90\% of parameters (passage from $N=128$ to $N=32$), the final values are always higher by less than 4\%, for both models. The different final values of the loss function can be represented on scatter plots, see Figure \ref{fig:comparisonLossesTL2}. The latter clearly illustrates the robustness of the fixed-kinetic model, see Figure \ref{fig:comparisonLossesHTL2}. It also shows that models with $L=1$ perform poorer than those with $L=2, \ 10, \ 50$.

\subsection{Additional results on Encoder-free NHF models}\label{subsec:EncoderFree}

\begin{figure}[!ht]
\centering
 \includegraphics[width=0.4\hsize]{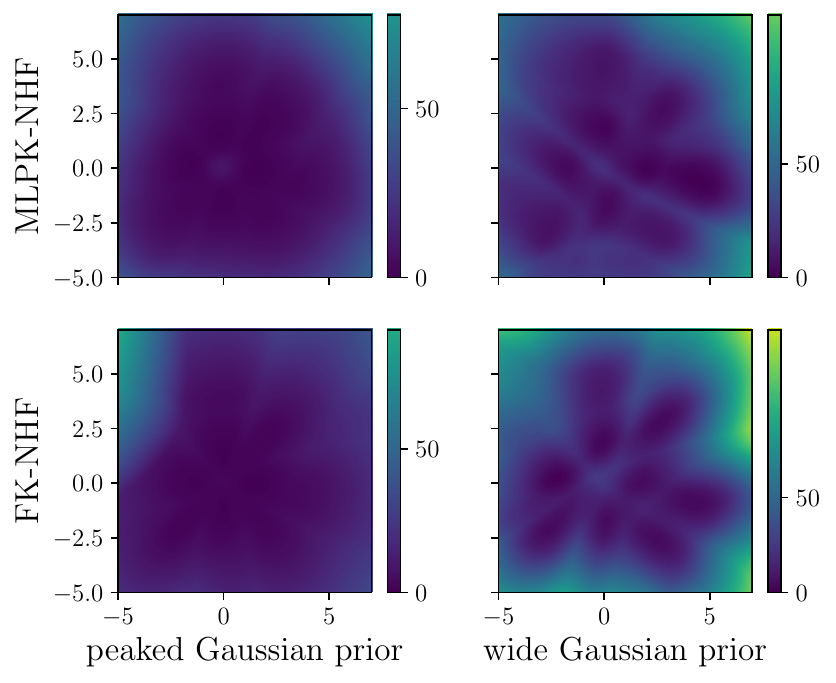} 
\caption{Shifted potential energy learned by 4 different Encoder-free models. All of them have recovered the 9 correct modes of the data.}
\label{fig:VwithoutEncoder}
\end{figure}

\begin{figure}[!ht]
\centering
 \includegraphics[width=0.4\hsize]{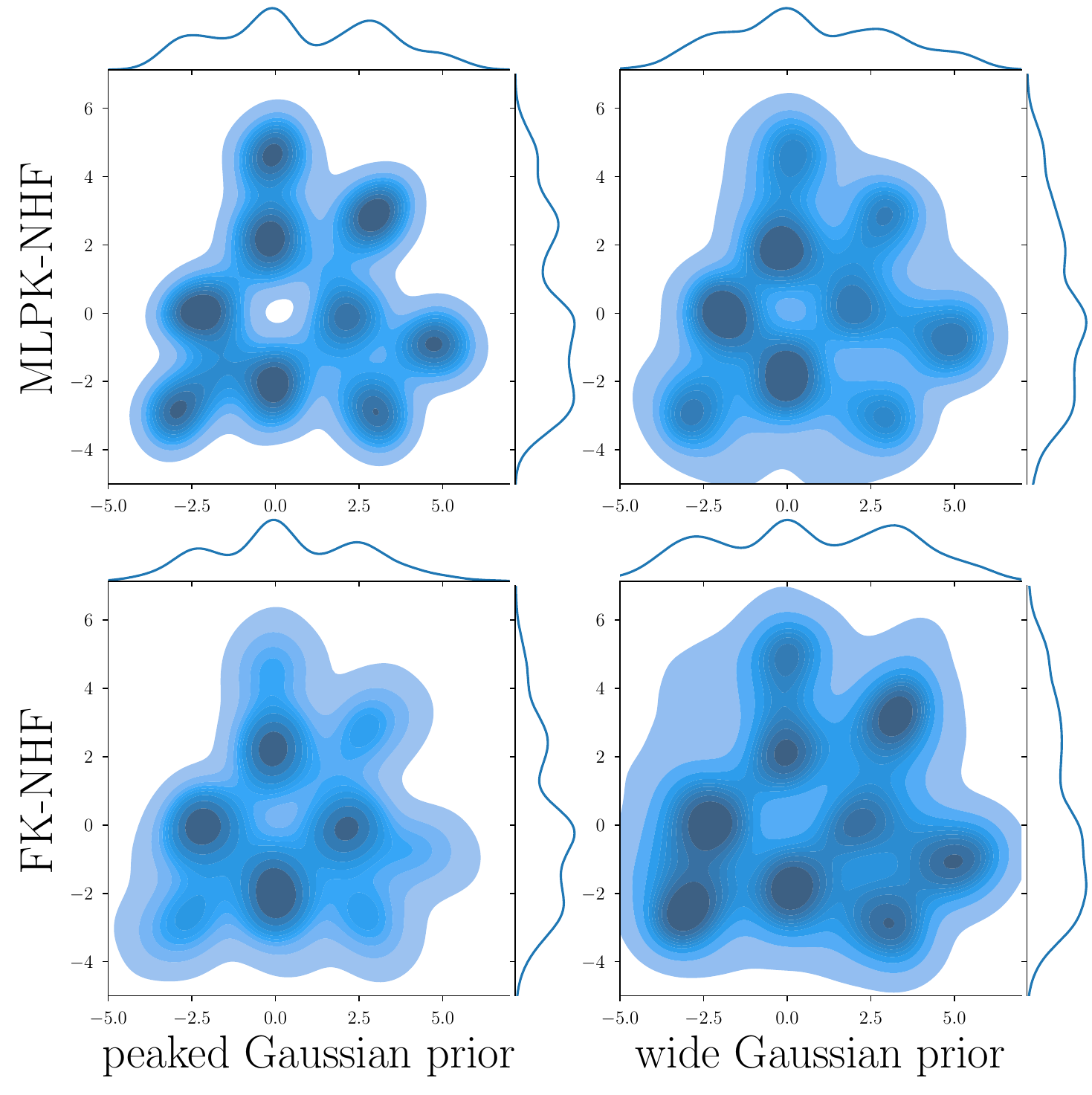} 
\caption{Density estimation of samples generated by 4 different Encoder-free models.}
\label{fig:qTwithoutEncoder}
\end{figure}

We tested different Encoder-free NHF models on the 2D Gaussian mixture target. As can be seen in Figure~\ref{fig:VwithoutEncoder}, all models have learned an interpretable multimodal potential. The corresponding samples are presented in Figure~\ref{fig:qTwithoutEncoder}. These experiments clearly illustrate that the transfer of multimodality and thus interpretability of the model can be achieved more easily by fixing the distribution of the artificial momenta $\mathbf{p}_T$ to a unimodal Gaussian. However, we chose to push the investigation further only with Encoder-based models as leaving as much flexibility as possible in the generation of momenta may be useful in sampling more challenging target distributions.

\section{NUMERICAL EXPERIMENTS ON THE MNIST DATASETS}\label{app:MNIST}

\subsection{Experimental details}\label{subsec:expDetailsMNIST}

The considered models were tested on the MNIST handwritten digits and Fashion MNIST datasets, which contain 60,000 images of size $28 \times 28$. The energy functions within the MLP-kinetic and Fixed-kinetic NHF are parameterized by 3-hidden layer MLPs with size $(784,512,256,128,1)$. As for $\mu$ and $\sigma$, they are 3-hidden layer MLPs with size $(784,256,256,256,784)$. We use LeakyReLU activation functions in between hidden layers with slope $0.1$ for $\mu$ and $\sigma$ and Softplus activation functions in between hidden layers for the energies. For the Fixed-Kinetic model, the mass matrix is optimized on the fly during training by learning its Cholesky decomposition, which has $D(D+1)/2$ parameters, $D$ being the dimension of data, as was done in \cite{CELLEDONI2023114608}. This represents a grand total of $1.94$ million learnable parameters for Fixed-kinetic NHF and $2.20$ million for MLP-Kinetic NHF, that both use $10$ Leapfrog steps with integration timestep $dt=0.1$. As for the Real NVP to which they are compared, it uses $6$ coupling layers and $32$ planes for a total of $2.27$ million learnable parameters. We adapted an architecture from a public open source GitHub project available at \url{https://github.com/bjlkeng/sandbox/tree/master/realnvp} under MIT Licence. All models were trained using a $\mathcal{N}(0,I_{784})$ base distribution, except for the NHF models on the MNIST handwritten digits which use a $\mathcal{N}(0,2^2I_{784})$. All models were trained for $50$ epochs on minibatches with size $32$ and optimization was performed using the Adam algorithm \cite{Adam}. The experiments were run on a HPC cluster, each of them using one GPU.

\subsection{Additional quantitative results}
\label{subsec:hist_bit}

\begin{figure}[!ht]
    \centering
    \includegraphics[scale=0.5]{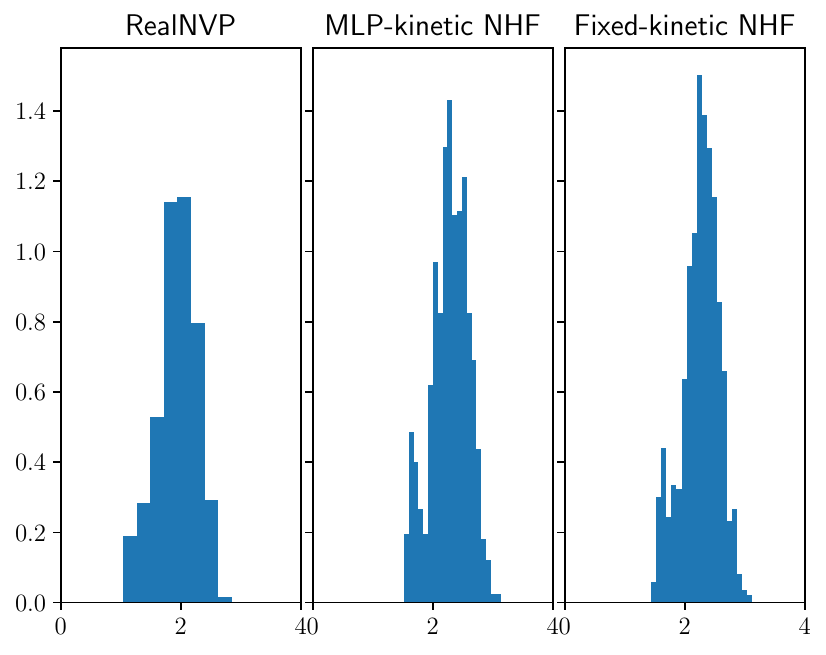}
    \caption{Histogram of bits per pixel values from the MNIST handwritten digits dataset. The standard deviation is approximately equal to $0.3$ for each model.}
    \label{fig:digitsbppHisto}
\end{figure}

\begin{figure}[!ht]
    \centering
    \includegraphics[scale=0.5]{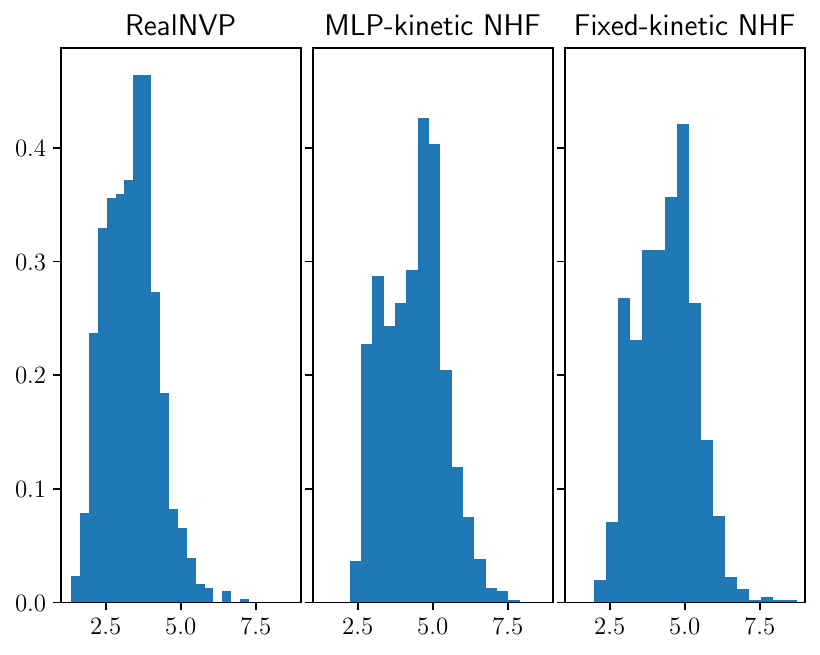}
    \caption{Histogram of bits per pixel values from the Fashion MNIST dataset. The standard deviation is approximately equal to $1$ for each model.}
    \label{fig:fashionbppHisto}
\end{figure}

It should be noted that the reported values of bits per pixel for MNIST handwritten digits and Fashion MNIST correspond to an average over some samples drawn from the true test dataset. In Figure~\ref{fig:digitsbppHisto} and Figure~\ref{fig:fashionbppHisto} are plotted the histograms of the bits per pixel values for $1024$ samples drawn from the true test datasets, for each model. Figure \ref{fig:digitsbppHisto} on the MNIST dataset exhibits comparable results for the three tested architectures whereas the Fashion MNIST experiment shows slightly better bits per pixel values for RealNVP.

\subsection{Interpretability of NHF models in higher dimension}\label{subsec:InterpretabilityMNIST}

\begin{figure}[!ht]
    \centering
    \includegraphics[width=0.6\hsize]{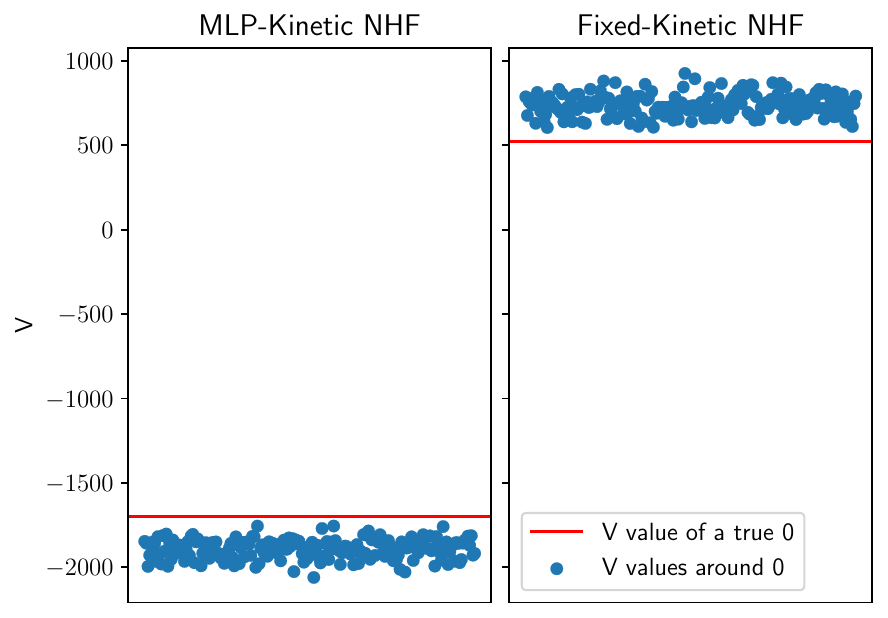}
    \caption{For two NHF models trained on the MNIST handwritten digits, we plot the value of the potential for a true '0' corresponding to a point $x_0$ in a $784$-dimensional logit space, as well as the values of the potential for 200 perturbations of the form $x=x_0+\varepsilon$ where $\varepsilon \sim \mathcal{N}(0,2^2I_{784})$.}
    \label{fig:potentialsInHD}
\end{figure}

Both NHF models preserve their interpretability properties even in high dimension image generation problems. They have learned extrema at the modes of data. As can be seen in Figure~\ref{fig:potentialsInHD}, the fixed-kinetic model has learned local minima at the modes of data while the usual MLP-kinetic model has learned local maxima. Imposing a positive quadratic term for the kinetic energy ensures that it will always be minima. 

\begin{figure}[!ht]
    \centering
    \includegraphics[width=0.6\hsize]{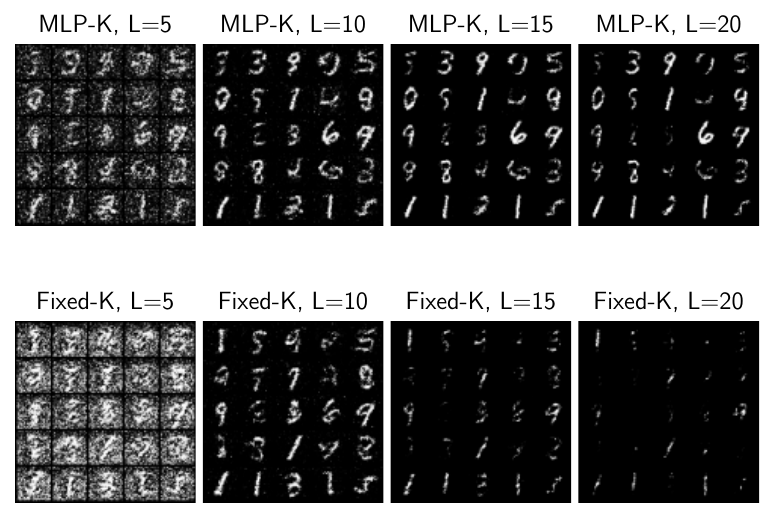}
    \caption{Stability of both NHF models trained with $L=10$ to the number of Leapfrog steps, for the MNIST handwritten digits dataset.}
    \label{fig:stabilityMNIST}
\end{figure}

\begin{figure}[!ht]
    \centering
    \includegraphics[width=0.6\hsize]{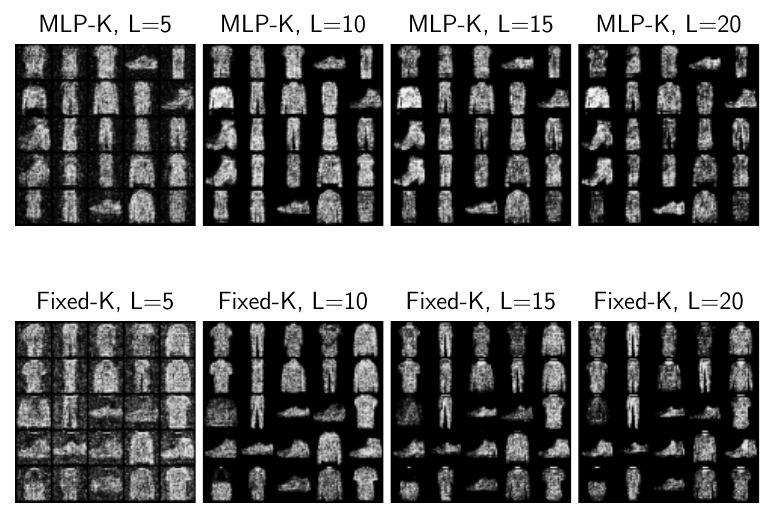}
    \caption{Stability of both NHF models trained with $L=10$ to the number of Leapfrog steps, for the Fashion MNIST dataset.}
    \label{fig:stabilityFashionMNIST}
\end{figure}

Again, by learning an interpretable potential we observe that the model is less sensitive to the number of Leapfrog steps. This is particularly visible for MLP-kinetic NHF on both MNIST handwritten digits and Fashion MNIST datasets, as well as for FK-NHF especially on Fashion MNIST, see Figure~\ref{fig:stabilityMNIST} and Figure~\ref{fig:stabilityFashionMNIST}.

\section{DERIVATION OF THE KL-DIVERGENCE AND ELBO FOR THE INFERENCE PROBLEM}
\label{app:lossesInference}

The KL-divergence suited to the inference problem is derived as follows:

\begin{scriptsize}\begin{equation}\label{KL2}
\begin{split}
    &D_{KL}(M(\mathbf{q}_T,\mathbf{p}_T) \ || \ \pi(\mathbf{q}_T|\boldsymbol{d}) g(\mathbf{p}_T)) = \int M(\mathbf{q}_T,\mathbf{p}_T) \log M(\mathbf{q}_T,\mathbf{p}_T) d\mathbf{q}_Td\mathbf{p}_T - \int M(\mathbf{q}_T,\mathbf{p}_T) \left[\log \pi(\mathbf{q}_T|\boldsymbol{d}) + \log g(\mathbf{p}_T) \right] d\mathbf{q}_Td\mathbf{p}_T \\
    &= \int \Pi_0(\mathcal{T}^{-1}(\mathbf{q}_T,\mathbf{p}_T)) \log \Pi_0(\mathcal{T}^{-1}(\mathbf{q}_T,\mathbf{p}_T)) d\mathbf{q}_Td\mathbf{p}_T - \int M(\mathbf{q}_T,\mathbf{p}_T) \left[ \log \pi_0(\mathbf{q}_T) + \log \ell(\boldsymbol{d}|\mathbf{q}_T) - \log p(\boldsymbol{d}) + \log g(\mathbf{p}_T) \right] d\mathbf{q}_Td\mathbf{p}_T \\
    &= \int \Pi_0(\mathbf{q}_0,\mathbf{p}_0) \left[ \log \pi_0(\mathbf{q}_0) +\log f(\mathbf{p}_0|\mathbf{q}_0) \right] d\mathbf{q}_0d\mathbf{p}_0 - \int M(\mathbf{q}_T,\mathbf{p}_T) \left[ \log \pi_0(\mathbf{q}_T) + \log \ell(\boldsymbol{d}|\mathbf{q}_T) + \log g(\mathbf{p}) \right] d\mathbf{q}_Td\mathbf{p}_T + \text{cst} \\
    &= \int \Pi_0(\mathbf{q}_0,\mathbf{p}_0) \left[\log \pi_0(\mathbf{q}_0) +\log f(\mathbf{p}_0|\mathbf{q}_0) - \log \pi_0(\mathcal{T}_q(\mathbf{q}_0,\mathbf{p}_0)) - \log \ell(\boldsymbol{d} | \mathcal{T}_q(\mathbf{q}_0,\mathbf{p}_0)) - \log g(\mathcal{T}_p(\mathbf{q}_0,\mathbf{p}_0)\right] d\mathbf{q}_0d\mathbf{p}_0 + \text{cst}.
\end{split}
\end{equation}\end{scriptsize}

We can also adapt the ELBO to the inference framework:
\begin{align*}\label{inferenceELBO}
	\ln \pi_0(\mathbf{q}_0) &= \ln \int \Pi_0(\mathbf{q}_0, \mathbf{p}_0) d\mathbf{p}_0 \\
	&= \ln \int \frac{\Pi_0(\mathbf{q}_0,\mathbf{p}_0)}{f(\mathbf{p}_0|\mathbf{q}_0)} f(\mathbf{p}_0|\mathbf{q}_0) d\mathbf{p}_0 \\
	&= \ln E_f \left[ \frac{\Pi_0(\mathbf{q}_0,\mathbf{p}_0)}{f(\mathbf{p}_0|\mathbf{q}_0)} \right] \\
	&\geq E_f  \left[ \ln \Pi_0(\mathbf{q}_0,\mathbf{p}_0) - \ln f(\mathbf{p}_0|\mathbf{q}_0) \right] \\
	&= E_f  \left[ \ln M(\mathcal{T}(\mathbf{q}_0,\mathbf{p}_0)) - \ln f(\mathbf{p}_0|\mathbf{q}_0) \right].
\end{align*}
 Then, expliciting $M(q,p) = \pi_0(q)\ell(d|q)g(p)$:
\begin{align}
\text{ELBO}(\mathbf{q}_0)
  &= E_f  \left[ \ln \left[ \pi_0(T_q(\mathcal{T}(\mathbf{q}_0,\mathbf{p}_0)))\ell(d| T_q(\mathcal{T}(\mathbf{q}_0,\mathbf{p}_0)))g(T_p
  (\mathcal{T}(\mathbf{q}_0,\mathbf{p}_0))) \right] - \ln f(\mathbf{p}_0|\mathbf{q}_0) \right].
\end{align}

\section{SOME TECHNICAL DETAILS ABOUT THE COSMOLOGICAL PROBLEM}\label{app:CosmoDetails}

In Section 7, we defined a relationship between the distance modulus $\mu$ and the redshift $z$ which also depends on two cosmological parameters $\Omega_m$ and $h$. To be more specific:

$$	\mu(z,\Omega_m,h) = 5\log_{10}\left(\frac{D_L^*(z,\Omega_m)}{h 10\text{pc}}\right) 
$$
where
$$D_L^*(z,\Omega_m) = \frac{c(1+z)}{H_0} \int_{0}^{z} \frac{ds}{\sqrt{1-\Omega_m + \Omega_m(1+s)^3}},
$$
and $H_0=100~\text{km s}^{-1}~\text{Mpc}^{-1}$,
$c$ being the speed of light. 

In practice, we avoid computing the integral in $D^*_L$ by using an approximation from \cite{Pen_1999} which is only valid for a flat Universe:
$$    D_L^*(z,\Omega_m) = \frac{c(1+z)}{H_0} \left[ \eta(1,\Omega_m) - \eta\left(\frac{1}{1+z},\Omega_m\right) \right],
$$
with 
$$\eta(a,\Omega_m) = 2\sqrt{1+s^3} \left( \frac{1}{a} - 0.1540\frac{s}{a^3} + 0.4304\frac{s^2}{a^2} + 0.19097\frac{s^3}{a} + 0.066941s^4 \right).$$

Note that the formal definition of these quantities imposes constraints on the possible values of the parameters, that can only be comprised between zero and one. We avoid the problem by outputting a sigmoid of $\mathbf{q}_T$.

\end{appendix}

\end{document}